\documentclass[conference]{IEEEtran}
\usepackage{cite}
\usepackage{amsmath,amssymb,amsfonts}
\usepackage{algorithmic}
\usepackage{graphicx}
\usepackage{textcomp}
\usepackage{xcolor}
\usepackage{hyperref}
\usepackage{url}
\usepackage{booktabs} 
\usepackage{url} 
\usepackage[numbers]{natbib}

\def\BibTeX{{\rm B\kern-.05em{\sc i\kern-.025em b}\kern-.08em
    T\kern-.1667em\lower.7ex\hbox{E}\kern-.125emX}}
\begin{document}

\title{Zero-Shot Visual Reasoning by Vision-Language Models: Benchmarking and Analysis}

\author{\IEEEauthorblockN{Aishik Nagar}
\IEEEauthorblockA{\textit{ASUS Intelligent Cloud Services (AICS)} \\
\textit{ASUS Global Pte. Ltd.}\\
Singapore, Singapore \\
aishiknagar@gmail.com}
\and
\IEEEauthorblockN{Shantanu Jaiswal \& Cheston Tan}
\IEEEauthorblockA{\textit{Centre for Frontier AI Research (CFAR)} \\
\textit{Institute of High Performance Computing (IHPC)}\\
\textit{Agency for Science, Technology and Research (A*STAR)}\\
jaiswals\_shantanu@ihpc.a-star.edu.sg, cheston\_tan@cfar.a-star.edu.sg}
}

\maketitle

\begin{abstract}
Vision-language models (VLMs) have shown impressive zero- and few-shot performance on real-world visual question answering (VQA) benchmarks, alluding to their capabilities as visual reasoning engines. However, the benchmarks being used conflate ``pure'' visual reasoning with world knowledge, and also have questions that involve a limited number of reasoning steps. Thus, it remains unclear whether a VLM's apparent visual reasoning performance is due to its world knowledge, or due to actual \textit{visual} reasoning capabilities.  

To clarify this ambiguity, we systematically benchmark and dissect the zero-shot visual reasoning capabilities of VLMs through synthetic datasets that require minimal world knowledge, and allow for analysis over a broad range of reasoning steps. 
We focus on two novel aspects of zero-shot visual reasoning: i) evaluating the impact of conveying scene information as either visual embeddings or purely textual scene descriptions to the underlying large language model (LLM) of the VLM, and ii) comparing the effectiveness of chain-of-thought prompting to standard prompting for zero-shot visual reasoning.

We find that the underlying LLMs, when provided textual scene descriptions, consistently perform better compared to being provided visual embeddings. In particular, $\sim$18\% higher accuracy is achieved on the PTR dataset. We also find that CoT prompting performs marginally better than standard prompting only for the comparatively large GPT-3.5-Turbo (175B) model, and does worse for smaller-scale models. This suggests the emergence of CoT abilities for visual reasoning in LLMs at larger scales even when world knowledge is limited. Overall, we find limitations in the abilities of VLMs and LLMs for more complex visual reasoning, and highlight the important role that LLMs can play in visual reasoning. 
\end{abstract}


\section{Introduction}
The development of vision-language models or VLMs \citep{tan2019lxmert,li2020oscar,wang2022image,alayrac2022flamingo,li2023blip,liu2023visual} has gained considerable attention in recent years given their application in developing general-purpose multimodal intelligence. Similar to the zero-shot abilities observed in large language models or LLMs \citep{brown2020language,chung2022scaling,chowdhery2022palm,touvron2023llama} for language tasks, VLMs such as Flamingo \citep{alayrac2022flamingo} and BLIP-2 \citep{li2023blip} have shown impressive zero- or few-shot reasoning abilities for language-vision tasks. Notably, they have been shown to surpass task-specific state-of-the-art models \citep{alayrac2022flamingo,li2023blip} when finetuned on common visual question answering (VQA) benchmarks, including VQAv2 \citep{goyal2017making} and OK-VQA \citep{marino2019ok}. Furthermore, recent works \citep{lu2022learn,zhang2023multimodal} have also shown how multimodal chain-of-thought (CoT) reasoning, wherein both language and vision modalities are used to elicit multi-step inference, improves the performance of models on multimodal question answering benchmarks such as ScienceQA \citep{lu2022learn}. These findings suggest that similar to LLMs, with increases in model size \citep{wei2022emergent} and advanced prompting techniques \citep{weichain,kojimalarge,jin2022good}, VLMs can exhibit stronger reasoning capabilities and operate as instruction-prompted zero- or few-shot visual reasoning engines. 

However, the current VQA benchmarks \citep{goyal2017making,marino2019ok,hudson2019gqa} used to evaluate the visual reasoning abilities of VLMs predominantly contain questions requiring only a few reasoning steps, and they often conflate visual reasoning with factual or world knowledge. 
While open-world visual reasoning certainly relies on knowledge of the world, it is important to recognize that \textit{visual} reasoning at its core encompasses a wide range of cognitive processes including scene interpretation, memory manipulation, spatial reasoning or attention, and logical or semantic inference. To illustrate the above points further, consider the example question ``Who is wearing glasses?'' (given an image of two individuals) in the popular VQAv2 benchmark. A VLM's accurate answer to this question may simply be due to world knowledge about ``glasses'' and different categories of ``persons'', and not necessarily due to better \textit{visual} reasoning capabilities. Similarly, the OK-VQA dataset is particularly designed to test how well models can utilize general world knowledge for VQA, and contains questions such as ``What phylum does this animal belong to?'' (given an animal image).

As such, based on the evaluation benchmarks and analysis in existing works, it is uncertain whether a model's apparent visual reasoning performance is due to its knowledge of the world, or its actual \textit{visual} reasoning capabilities. 

Thus, in this work, we propose to systematically analyze and benchmark zero-shot visual reasoning capabilities of VLMs through the usage of synthetic datasets. Specifically, we utilize the CLEVR \citep{johnson2017clevr} and PTR \citep{hong2021ptr} datasets, which contain questions requiring minimal world knowledge, but a broader range of ``reasoning steps'' and primitive visual reasoning operations. Moreover, these datasets provide detailed meta-information for each (question, image) pair, including a complete symbolic scene description, as well as a step-by-step functional program for the question. Cumulatively, the broader range of complexities and associated meta-information allow us to better quantify and draw conclusions regarding the ``pure'' visual reasoning capabilities of VLMs. Additionally, they enable us to assess performance across different fundamental visual operations such as counting, attribute or relationship detection and physical or analogical inferences. 

\subsection{Summary of Experiments and Findings}

We focus on investigating two novel aspects of zero-shot visual reasoning in VLMs. Firstly, we compare the performances of VLMs versus LLMs. Specifically, we compare a ``traditional VLM'' (i.e. an LLM receiving scene information as visual embeddings from a base vision model) against an LLM simply receiving a completely textual representation of the scene. We find that LLMs consistently outperform VLMs that utilize the same base LLMs. Specifically, in the case of the BLIP2-Flan-T5 \citep{li2023blip} model, using only its base LLM, i.e. Flan-T5 \citep{chung2022scaling}, without the visual front-end achieves $\sim$18\%  higher accuracy on the PTR dataset, while GPT-4 was $\sim$17\% more accurate than GPT-4V on CLEVR. One key takeaway is that for questions which can be solved in 2 to 5 ``reasoning steps'', LLMs show performance levels which are significantly above chance, suggesting that LLMs may in fact possess reasonable capabilities as zero-shot visual reasoning engines.

Secondly, we study how CoT prompting compares to standard prompting for zero-shot application of these models in the context of VQA. We find that CoT prompting for visual reasoning in LLMs only obtains better results than standard prompting at large model scales (in our case for the 175B GPT-3 turbo model) and performs worse for smaller models. For LLMs and VLMs, we observe trends of emergence of CoT reasoning in zero shot settings even when the model's knowledge and context about the world is restricted. Furthermore, owing to the use of synthetic datasets to benchmark VLMs which are not explicitly trained on reasoning on synthetically rendered scenes, we also observe than increase model scale shows signs of improving CoT reasoning capabilities. This indicates that model scaling and CoT could potentially be used to extend and improve zero-shot reasoning performance for multimodal models on previously unseen settings.

\subsection{Contributions}

(1) To our knowledge, we are the first to systematically benchmark zero-shot visual reasoning capabilities of VLMs using synthetic datasets, disentangling the impact of world knowledge, so as to assess ``pure'' visual reasoning by models. 

(2) We compare the zero-shot VQA performance of VLMs against LLMs, and find that LLMs receiving only ground-truth textual scene information consistently perform better than when provided with visual embeddings.

(3) Consistent with previous studies on CoT for language tasks \citep{weichain}, we find CoT for visual reasoning in LLMs also seems to emerge for larger model sizes even when the model's world knowledge is limited.

(4) We analyze the visual reasoning performance of VLMs and LLMs under various factors including the number of ``reasoning steps'', question types and model scale. Our overall analysis indicates the limitations of VLMs and LLMs for complex visual reasoning and highlights the important role LLMs can play in enhancing visual reasoning capabilities.

\section{Related Work}
\textbf{Benchmarking reasoning abilities of LLMs and VLMs.} Since the initial demonstration of LLMs as being effective few-shot learners \citep{brown2020language}, multiple works \citep{brown2020language,chung2022scaling,ouyang2022training,jin2022good,touvron2023llama} have sought to refine the design and training of LLMs, besides comprehensively benchmarking \citep{liang2022holistic,srivastava2022beyond,valmeekam2022large} their reasoning abilities  on language-specific tasks. More recently, the development of VLMs \citep{tan2019lxmert,li2020oscar,wang2022image,alayrac2022flamingo,li2023blip,liu2023visual} has drawn on advancements in both LLMs and vision-foundation models leading to their prompt-based application for vision-language tasks \citep{alayrac2022flamingo,li2023blip,liu2023visual,wu2023visual} such as image captioning, text-guided image editing and general VQA. These works have evaluated the performances of VLMs on prominent VQA benchmarks including VQA-v2 \citep{goyal2017making}, OK-VQA \citep{marino2019ok}, GQA \citep{hudson2019gqa} and VizWiz~\citep{gurari2018vizwiz} in zero-shot, few-shot and fine-tuned settings. However, these analyses are not sufficient to conclude the ``true'' visual reasoning capabilities of VLMs since the datasets typically conflate world knowledge with visual reasoning and require only a limited number of ``reasoning steps'' \citep{nagar2024dissecting}. Further, these works have not assessed whether LLMs by themselves when provided textual scene representations are capable of visual reasoning. Thus, our work aims to more comprehensively evaluate the zero-shot visual reasoning capabilities of VLMs and their underlying LLMs by utilizing synthetic datasets.

\textbf{Synthetic datasets to disentangle reasoning capabilities from world knowledge}. There are several synthetic datasets which can disentangle world knowledge from reasoning in different ways. \citep{suhr2017corpus} is a dataset designed for visual reasoning tasks. The images are synthetic and often involve simple shapes and layouts, ensuring the focus is on reasoning rather than world knowledge. \citep{kuhnle2017shapeworld} generates abstract visual scenes and accompanying textual descriptions  designed to test various linguistic and visual phenomena. \citep{Zhang_2019_CVPR} is a synthetic visual reasoning dataset inspired by the structure of Raven's Progressive Matrices, a popular human IQ test. This format ensures that success on the task requires true visual reasoning and pattern recognition, rather than relying on learned associations or world knowledge. \citep{johnson2017clevr} and \citep{hong2021ptr}, the datasets used in this study, are tailored for disentangling world knowledge from visual reasoning. Their machine-generated questions ensure controlled complexity to test visual reasoning without relying on pre-trained visual or linguistic biases. Their rich annotations and scene metadata are ideal for testing reasoning abilities not only the visual and spatial aspects of the scene, but also the physical interactions in a wide range of reasoning types.

\textbf{CoT prompting for zero- or few-shot reasoning.}
The development of CoT techniques \citep{weichain,kojimalarge,yao2023tree}, wherein models are elicited to reason in multiple steps, has been shown to significantly benefit zero- or few-shot performance of LLMs on diverse language and logical reasoning tasks. \citep{kojimalarge} More recently, CoT techniques have been developed \citep{lu2022learn,zhang2023multimodal} to incorporate both vision and language modalities in finetuning LLMs for multimodal question-answering benchmarks such as ScienceQA \citep{lu2022learn}. In contrast to these works, we specifically analyze the impact of CoT prompting for zero-shot VQA.

\section{Experiments}
\label{experiments}
\subsection{Experimental Design}
Our experiment design philosophy was primarily guided by the major benchmarks and analysis which we wanted to perform in this study. 
Our first goal was to analyze the impact of scene information representation in the form of text or images on the model's zero-shot reasoning capabilities.  Based on this, we provided the complete scene information in text format to the LLM (the Flan-T5 model family) using the scene metadata, while providing the scene image to the model's VLM counterpart, which was the BLIP-2 Flan-T5 model family \citep{li2023blip}. To gauge the impact of the text-based scene metadata on VLM performance, we also ran a set of experiments providing both the scene metadata and the image to the VLM. Through this setup, we could study areas where the VLM might fall short in terms of information extraction and reasoning, and also identify if there were specific reasoning categories where direct visual representation might be a clear advantage. 
The second goal was to identify the impact of Chain-of-Thought prompting on the reasoning abilities of LLMs and VLMs as well as its performance trends over scale, when the models world knowledge is limited. To achieve this we designed experiments which could benchmark different scale models of the same LLM and their counterpart VLM families on CoT and Standard Prompts.

\subsection{Experimental Setup}
\begin{figure*}[htbp]
  \centering
    \includegraphics[width=\textwidth]{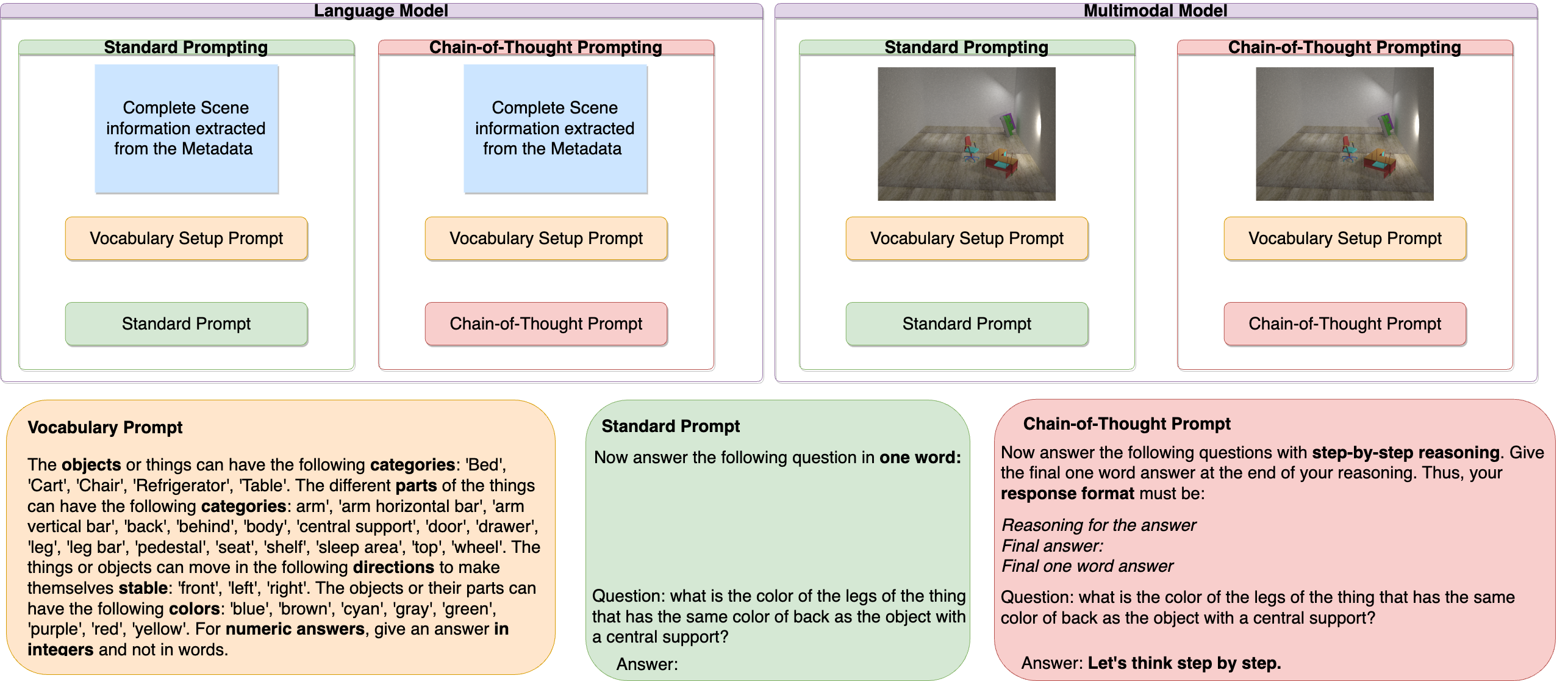}
   \caption{The experimental setup. We perform experiments on pure LLMs as well as their VLM variants with the same set of prompts. In case of LLMs, the image information is provided using the scene metadata used to render the image.}
   \label{fig_setup}
\end{figure*}

\textbf{Datasets}. We use two datasets: (1) CLEVR~\citep{johnson2017clevr}, a synthetic VQA dataset containing images of 3D-rendered objects; each image comes with a number of compositional questions of various types, and (2) PTR~\citep{hong2021ptr}, a dataset for part-based conceptual, relational and physical reasoning. Since the scene metadata was only provided for the images in the train and validating sets (and not the test sets), we use the validation sets of each of these datasets for testing. This allowed us to automatically generate text descriptions of the scenes to compare performance of Visual Language Models (VLMs) with the pure LLMs. There is neither training nor validation per se, since our experiments are in a zero-shot setting.

\textbf{Standard prompting}. Our standard prompting procedure included providing the models with the relevant scene information (the image in the case of VLMs, or the scene metadata in the case of pure LLMs), a setup prompt and instructing the model to provide the final answer directly in one word. Since the models were tested in a purely generative setting, the models would often generate the correct answer, but not use the correct terminology, e.g. calling a cyan object light blue. To maintain the generative setting but align the model answers to match the scene terminology, it was provided with the setup prompt, which gave information on the possible attributes, colors, shapes etc. which could be present.

\textbf{Chain-of-Thought Prompting}. To elicit CoT reasoning in a zero-shot setting, we follow the prompt template of~\citep{kojimalarge}. In addition to the same information and setup prompt provided in the standard prompt, we add ``Let's think step by step'' before each answer. We also developed a format prompt to force the model to give a final one-word answer at the end.

\textbf{Visual Language Models}. We used three VLMs tuned for instructed generation for the experiments. These are BLIP2-Flan-T5-XL (3B) and BLIP2-Flan-T5-XXL (11B) ~\citep{li2023blip} and GPT-4V. Using these models allowed us to compare the performance of the VLMs against the pure LLM versions of these models. Pretrained weights from LAVIS~\citep{li2022lavis} were used for the BLIP-2 Flan-T5 model family.
\textbf{Language Models}. For comparing VLMs to LLMs, we used three LLMs: Flan-T5-XL (3B), Flan-T5-XXL (11B)~\citep{chung2022scaling} and GPT-4. For CoT reasoning, such abilities have been shown to emerge at a scale of more than 100B~\citep{weichain}. Thus, we also tested our setup on GPT-3.5-Turbo (175B)~\citep{ouyang2022training} and smaller versions of GPT to analyse the impact of scale on CoT.

\section{Results and Analyses}

\subsection{Comparing LLMs with scene descriptions versus VLMs}
\begin{figure*}[htbp]
  \centering
    \includegraphics[width=\textwidth]{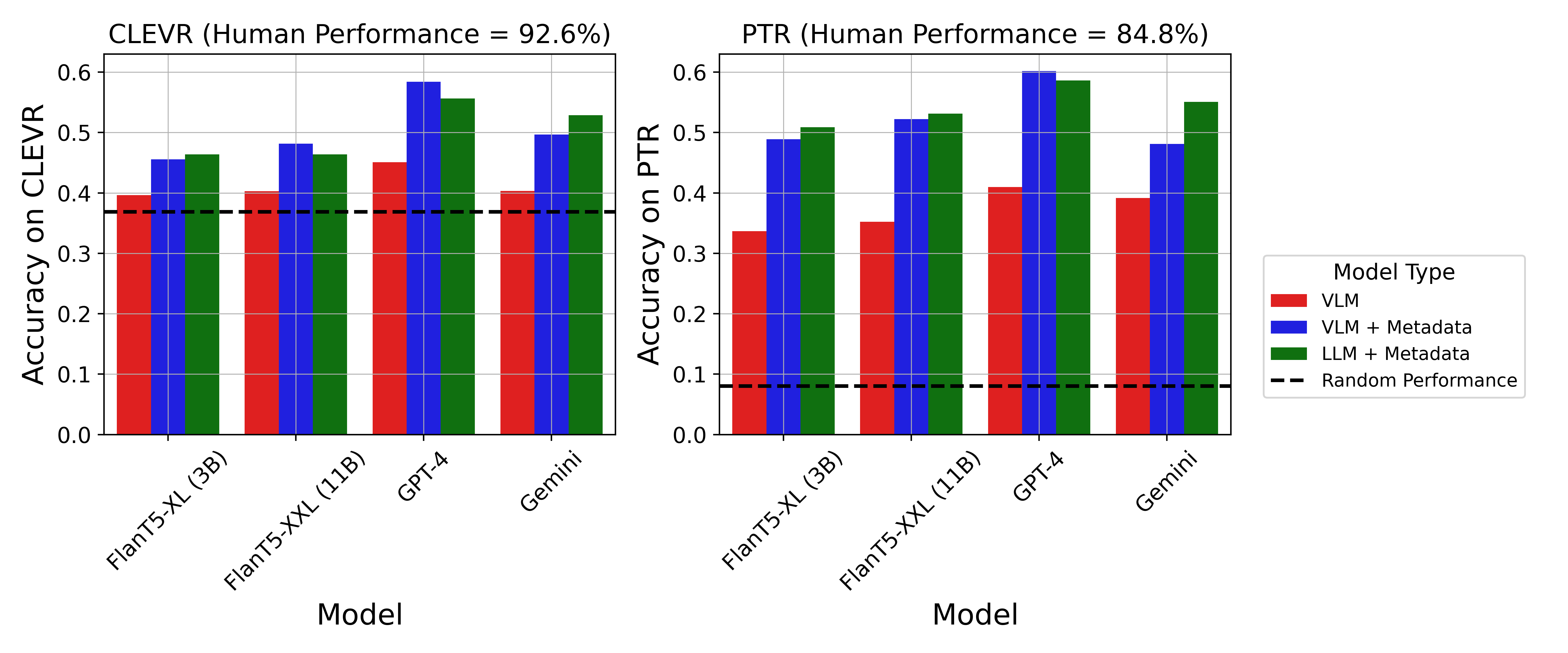}
  \caption{LLM versus VLM+Metadata versus VLM performance on CLEVR and PTR.}
  \label{fig:LLM-v-multimodal-MAIN}
\end{figure*}

{\textbf{LLMs with scene descriptions outperform VLMs}}: Figure~\ref{fig:LLM-v-multimodal-MAIN} shows the impact of visual grounding using BLIP-2 on the reasoning effectiveness of the models. Pure LLMs generally outperform or have similar performance to their counterpart VLM models across both scales and datasets. A t-test was performed to test if the pure LLMs performed better than VLMs. A p-value of 0.0088 indicates that the difference is statistically significant. This might seem counter-intuitive, as one might expect the VLM to be able to effectively utilize the ``visual frontend'' provided by the image encoder used in the BLIP-2 setup for querying the relevant aspects of the image. There are 2 possible explanations: 1) There are underlying issues in the VLM architecture which prevent the visual front-end from providing relevant information to the model. 2) The complexity of the tasks is not enough that a visual front-end which queries only the relevant information from the scene can be better than providing the complete, unfiltered information to the reasoning engine (the LLM).

\textbf{To guard against data contamination (i.e. LLMs exposed to CLEVR or PTR during training)}, we ran image-free baselines. Model performance was 36.85\% for CLEVR (chance = 36.86\%) and 10.16\% for PTR (chance = 8.03\%). The models performed at chance, indicating no contamination.

\textbf{LLM advantage for CLEVR versus PTR}:
The difference in performance between the LLM and the VLM is more pronounced in PTR than CLEVR. For CLEVR, the LLM outperforms the VLM by roughly 6-7\%, while for PTR the gap is roughly 17-18\%. One possible explanation is that the objects in PTR are more complex, with multiple parts, hence the task for the VLM's visual frontend is more challenging, and more errors and uncertainty are introduced. Providing the ground-truth scene description to the LLM eliminates this challenging visual frontend task. Conversely, the objects in CLEVR are simple geometric objects, hence access to the ground-truth scene description provides less of an advantage to the LLM.
\begin{figure*}[htbp]
  \centering
    \includegraphics[width=\textwidth]{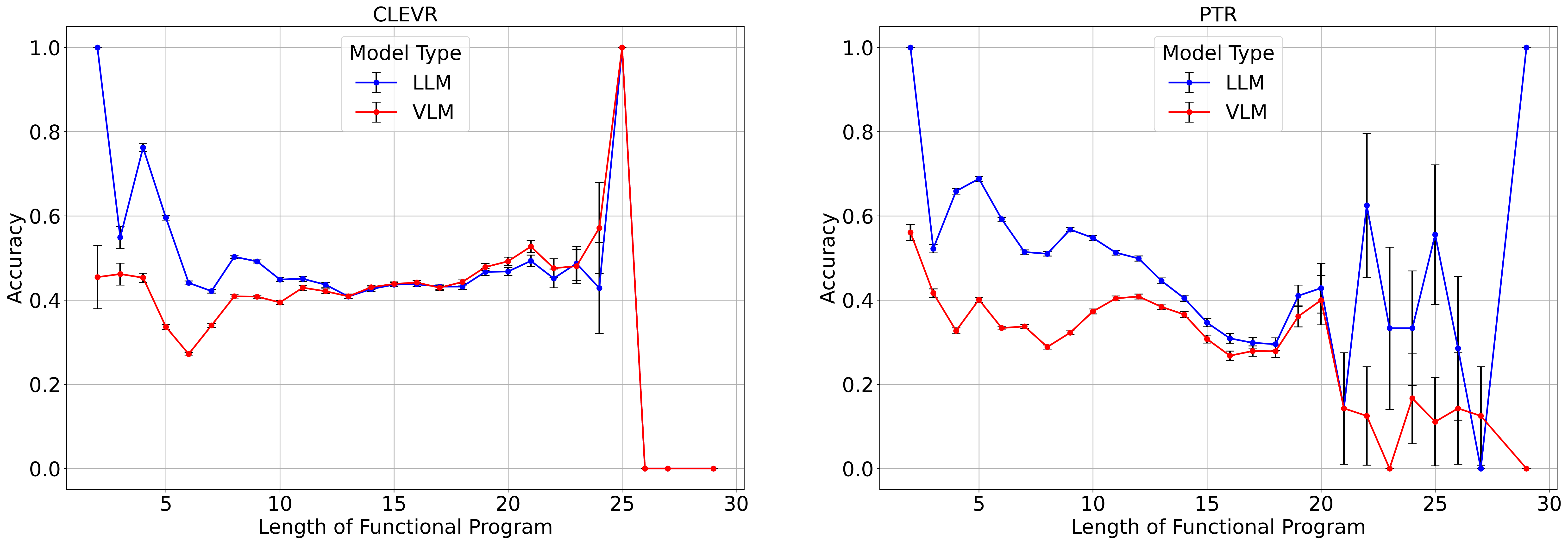}
  \caption{LLM versus VLM performance of Flan-T5-XXL on CLEVR and PTR, analyzed by length of functional programs (a proxy for number of reasoning steps). Error bars represent standard error; large error bars for functional programs longer than 18 are due to the small number of questions.}
    \label{fig:LLM-v-multimodal-STEPS}
\end{figure*}

\textbf{Analysis by number of ``reasoning steps''}: Both CLEVR and PTR provide functional programs which programmatically describe the solution for the reasoning tasks. We used the length of these functional programs \textbf{as a proxy} for the number of ``reasoning steps'' needed. We analyzed the results by number of ``reasoning steps'' (Fig.~\ref{fig:LLM-v-multimodal-STEPS}). For questions requiring relatively fewer ``reasoning steps'' (up to around 12-17), LLMs generally outperform VLMs. As seen in Fig.~\ref{fig:LLM-v-multimodal-STEPS} (right), for PTR, both LLMs and VLMs generally show declining performance as the number of ``reasoning steps'' increases, unsurprisingly. However, when it comes to CLEVR (Fig. \ref{fig:LLM-v-multimodal-STEPS}, left), the performance of VLMs seems to be somewhat independent of the number of ``reasoning steps''. This could be due to the nature of the CLEVR dataset; questions are usually abstract and require deep reasoning, regardless of the number of steps. As such, even tasks with fewer steps might demand similar levels of abstraction and reasoning as tasks with more steps.

Moreover, because CLEVR consists of geometric shapes rather than recognizable object parts, the VLMs may not gain as much valuable information from the visual encoder for each additional reasoning step. While the program length provides a heuristic for reasoning complexity, it might not always perfectly capture the cognitive complexity for humans. However, it is still worthwhile to study the impact of length of functional programs on performance.

\begin{figure*}[htbp]
  \centering
    \includegraphics[width=\textwidth]{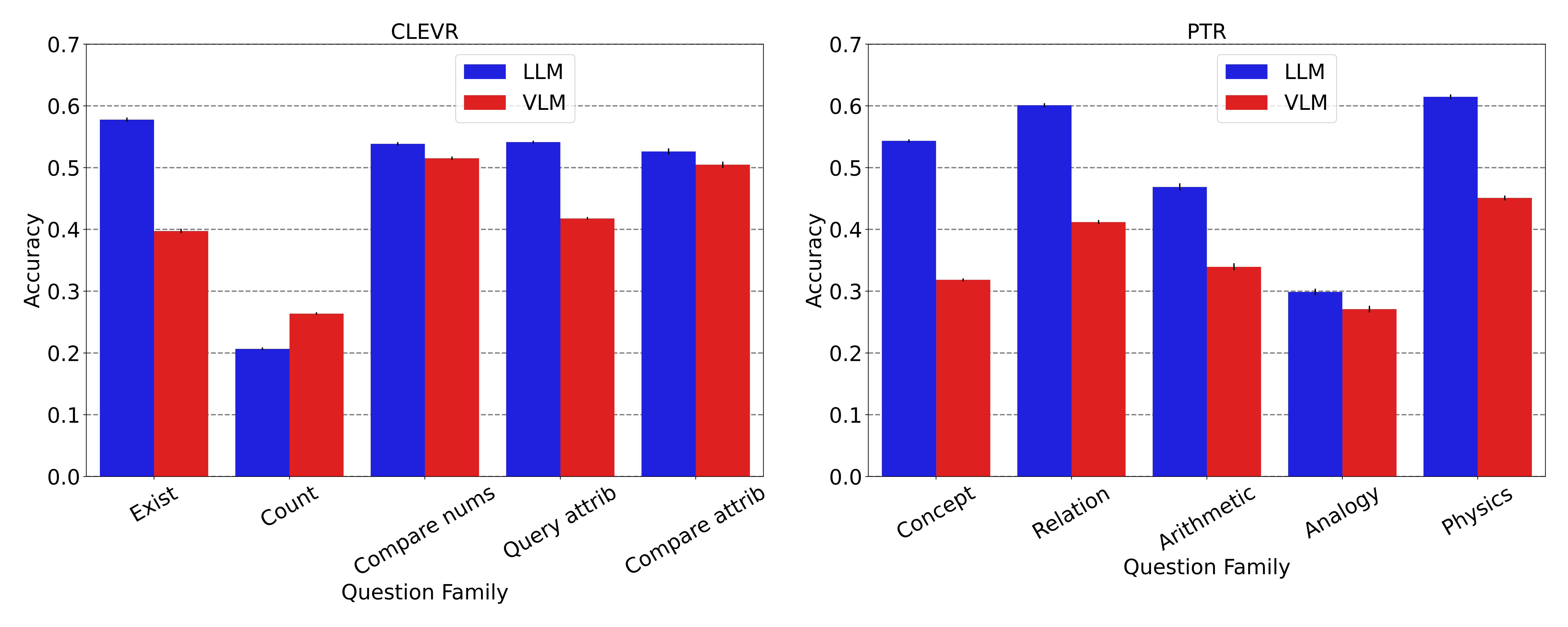}
  \caption{LLM versus VLM model performance of Flan-T5-XXL on CLEVR and PTR using standard prompting, organized by question family.}
  \label{fig:LLM-v-multimodal-QUESTIONS}
\end{figure*}

\begin{figure*}[htbp]
  \centering
    \includegraphics[width=\textwidth]{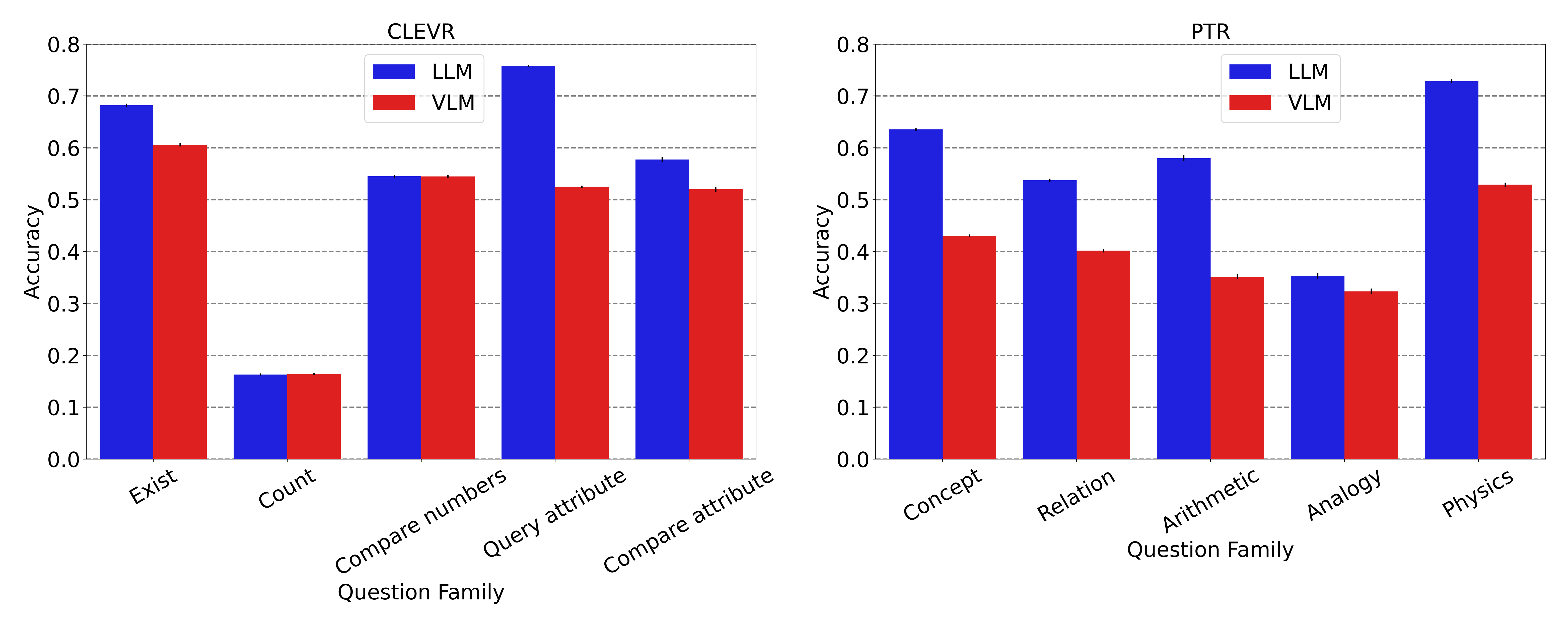}
  \caption{LLM versus VLM model performance of GPT-4 on CLEVR and PTR using standard prompting, organized by question family.}
  \label{fig:GPT-4-LLM-v-multimodal-QUESTIONS}
\end{figure*}

\textbf{Analysis by question family (CLEVR)}: The LLM performs better than the VLM in most categories (Fig.~\ref{fig:LLM-v-multimodal-QUESTIONS}, left and Fig. \ref{fig:GPT-4-LLM-v-multimodal-QUESTIONS}, left). The ``exist'' and ``query attribute'' categories show the largest differences in performance, with the LLM noticeably better. Interestingly, the VLM performs better in the ``count'' category for Flan-T5, while it is comparable to the LLM for GPT-4. The results could be explained by a few factors. For the LLMs, the ``exist'' and ``query attribute'' questions are the most straightforward tasks since this information just requires a direct lookup from the scene metadata. The VLMs, however, require identification of the correct object(s) and their attributes. For ``counting'' questions, on the other hand, it's possible that VLMs are more effective in tasks like counting where visual cues can be valuable. 

\textbf{Analysis by question family (PTR)}: The LLM outperforms the VLM across all question families on PTR (Fig.~\ref{fig:LLM-v-multimodal-QUESTIONS}, right and Fig. \ref{fig:GPT-4-LLM-v-multimodal-QUESTIONS}, right). The largest performance gap is observed in the ``concept'' and ``relation'' categories. ``Concept'' questions in PTR evaluate reasoning about basic part-whole relations. Similar to findings for CLEVR, question families which require simple ``lookups'' from the metadata have the largest gap in performance. Interestingly, the performance of LLMs on ``arithmetic'' questions is better than VLMs for this dataset (unlike the ``count'' questions in CLEVR). This can be attributed to the higher level of reasoning required for arithmetic questions. While such questions in CLEVR were limited to counting objects or comparing numbers, PTR questions require complex selection of object parts before performing arithmetic.

Visual analogy questions in PTR require complex reasoning that pose significant challenges for both LLMs and VLMs. This is evident from both having their worst performance on the ``analogy'' question family.  This process involves multiple stages, including identifying the relevant relationship, applying it to a new context, and generating or selecting the correct answer. The models must not only identify the relationship between A and B, but also accurately project it onto C and D. This complexity could make these tasks particularly challenging for both types of models. 
A key observation is that the VLM when provided with the scene metadata in addition to the image performed $\sim$2\% better than the base LLM only in the case of GPT-4, but not BLIP2. This indicates that the visual frontend  for GPT-4 provided additional benefits to the LLM for visual reasoning.

\textbf{Drawbacks of current VLM Architecture}: VLMs, even those leveraging LLMs, have architectural bottlenecks that may hinder performance. During inference, they function in two phases: 1) visual information querying, where the model's visual frontend extracts scene details based on a text query, and 2) text generation, where the LLM uses this extracted information for reasoning. This process lacks a feedback loop, preventing the LLM from requesting additional information if needed. In contrast, when LLMs receive full scene descriptions as text, they can access the entire description, thereby better retrieving relevant information. These drawbacks of VLM architectures are further evidenced by the fact that even when given access to scene metadata, VLMs consistently perform similar to LLMs. This suggests that they are unable to take advantage of the additional visual information.

\textbf{VLM performance on synthetic vs. real images.} One concern is that the VLMs are not trained on synthetic data, which could lead to lower performance compared to LLMs. We conducted experiments on the GQA~\citep{hudson2019gqa} dataset using a similar LLM vs. VLM comparison, and confirmed that the \textbf{LLMs also performed better than VLMs on natural images}. FLAN-T5 XXL performance was 78.72\%, while BLIP2 FLAN-T5 XXL performance was 56.81\%.

\subsection{Chain-of-Thought (CoT) Reasoning}
\begin{figure*}[htbp]
  \centering
    \includegraphics[width=\textwidth]{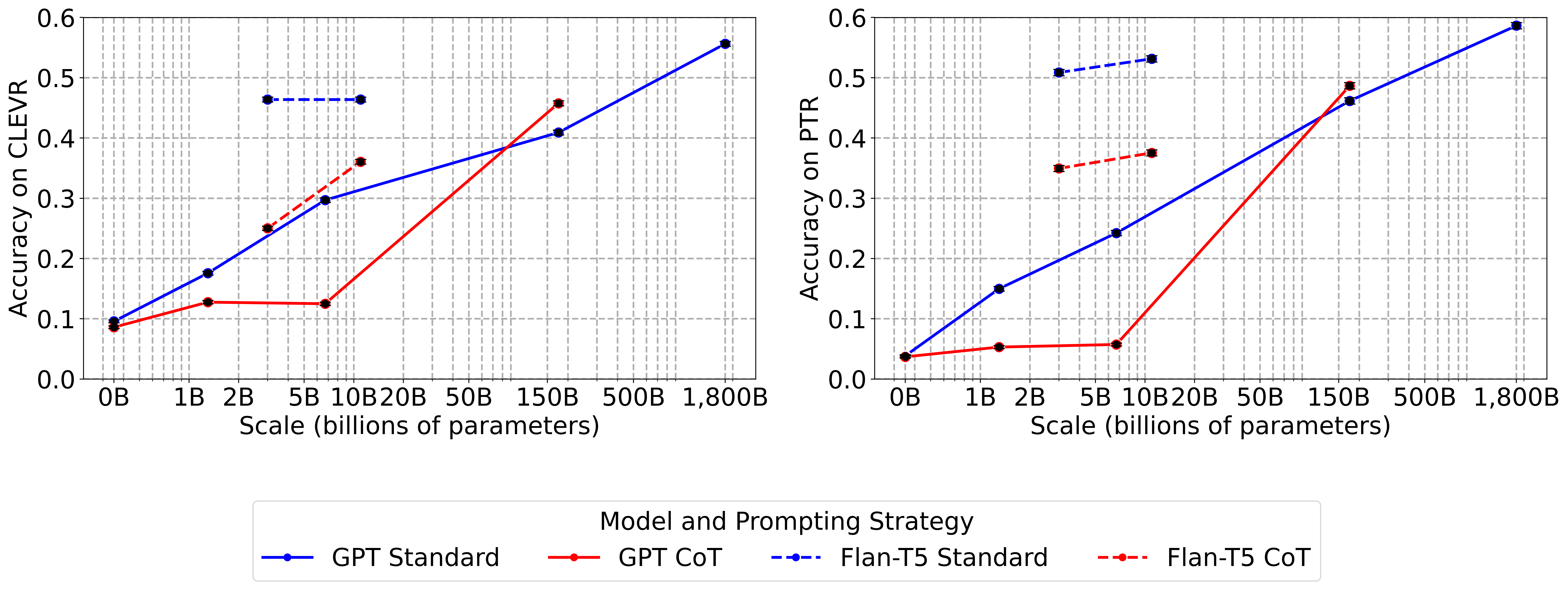}
   \caption{LLM performance on CLEVR and PTR datasets using standard and CoT prompting over scale. GPT-4 experiments were only performed for standard prompting due to cost constraints. The x-axis scale is logarithmic for better clarity.}
   \label{fig:CoT-MAIN}
\end{figure*}

\begin{figure*}[htbp]
  \centering
    \includegraphics[width=\textwidth]{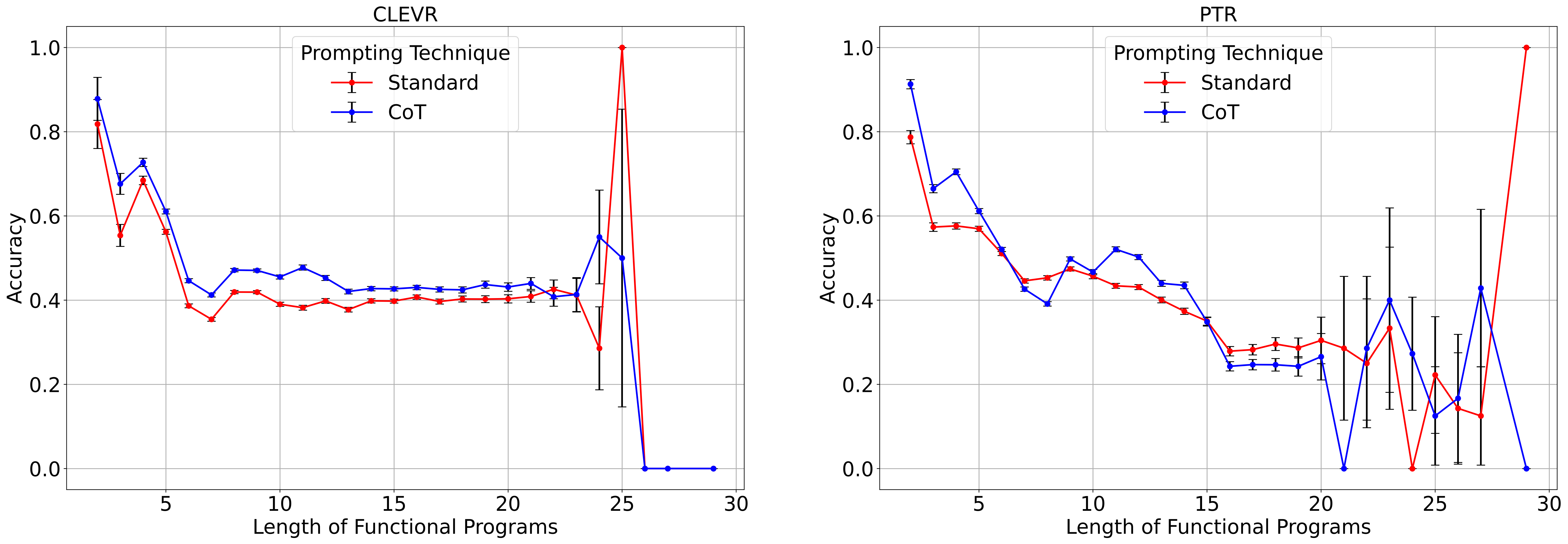}
  \caption{Standard versus CoT prompting performance of GPT-3.5-Turbo on CLEVR and PTR, analyzed by length of functional programs. The vertical black bars indicate standard error bars.}
  \label{fig:CoT-STEPS}
\end{figure*}

\textbf{Overall results}: Figure~\ref{fig:CoT-MAIN} presents a concise summary of the main outcomes of Chain-of-Thought reasoning on the two datasets. Interestingly, the open source Flan-T5-XXL (11B) model with standard prompting achieves the best performance, outperforming even GPT-3.5-Turbo (175B), which is over 15x larger. This is true for both datasets, and regardless of CoT or standard prompting for GPT-3.5-Turbo. Flan-T5-XL (3B) only performed marginally worse than its larger 11B cousin.

\textbf{Analysis by number of ``reasoning steps''}: As expected, performance generally drops with more ``reasoning steps'' (Fig.~\ref{fig:CoT-STEPS}). For CLEVR, CoT prompting produced a small but consistent performance gain over standard prompting. For PTR, the CoT advantage is less consistent, with standard prompting sometimes performing better.

\textbf{Analysis by question family (CLEVR)}: CoT prompting shows a noticeable improvement in the ``count'' question family, with some improvement in ``compare attribute'', ``compare numbers'' and ``exist'' categories. ``Query attribute'' questions in CLEVR typically involve direct queries about object properties, often solvable in a single step -- consistent with the fact that overall accuracy is highest for this question family. This could explain why CoT does not provide a significant advantage in this simple, often one-step question family.

\textbf{Analysis by question family (PTR)}: CoT prompting leads to improvements for ``relation'' and ``arithmetic'' questions. For ``analogy'' questions, CoT lowers performance. CoT assists in ``relation'' and ``arithmetic'' questions by breaking down the task into simpler steps, aiding in the understanding of sequential operations. On the other hand, for ``analogy'' questions, CoT might hinder performance by overly decomposing the problem, possibly losing sight of the overarching relationship.

\textbf{Impact of CoT across datasets}: CoT prompting resulted in significant improvements in the ``count'' category in CLEVR and ``arithmetic'' in PTR, both involving numerical understanding. A possible explanation is that these tasks are similar to text-based reasoning or step-by-step reasoning examples that LLMs may have encountered during training. However, the same degree of improvement was not observed in categories such as ``analogy'' and ``query attribute'', which are unique to visual reasoning and have no text-based equivalents.

The absence of significant improvement in visual reasoning tasks might be because base LLMs are not exposed to step-by-step visual reasoning samples during training. Consequently, CoT might not be effective for such tasks. This observation could also imply that the generalizability of CoT may be limited. Its effectiveness seems to be constrained to tasks that are similar to those encountered during training.
\textbf{CoT reasoning over scales:} From Fig.~\ref{fig:CoT-MAIN}, CoT prompting performs better than standard prompting only for the comparatively large GPT-3.5-Turbo (175B) model, and does worse for smaller models. These results suggest that the emergence of CoT reasoning is only at larger scales for visual reasoning, similar to prior observations for other reasoning tasks~\citep{weichain}.

\section{Conclusion}
We benchmarked zero-shot visual reasoning capabilities of VLMs and LLMs. We used synthetic datasets to mitigate the impact of world knowledge on visual reasoning performance, and to also evaluate reasoning over a broad range of ``reasoning steps''. We studied two novel aspects of zero-shot visual reasoning: i) evaluating how a VLM's base LLM performs when provided textual scene description, compared to when provided with a visual embedding, and ii) the effectiveness of CoT prompting. Further, we analyzed the performance of VLMs and LLMs under various factors, such as number of ``reasoning steps'', question types and model scale.

\section{Acknowledgement}
This work was supported by an A*STAR CRF award to CT.

\bibliographystyle{IEEEtranN}

\bibliography{new_references}

\appendix

\section{Appendix}
\subsection{Experiment Code and Reproducibility}
\label{code-details}

All the relevant code and scripts to process the dataset, run all experiments and evaluate the results is available with the supplemental submission . The code uses 2 major libraries for the experiments:

\begin{enumerate}
    \item \href{https://huggingface.co/docs/transformers/index}{The huggingface transformers library} for LLM experiments.
    \item \href{https://github.com/salesforce/LAVIS.git}{The Salesforce-LAVIS library} for VLM experiments.
\end{enumerate}

Setup instructions have been included in markdown where required. 

The 2 major datasets used (CLEVR, PTR and GQA), can be downloaded from these links:
\begin{enumerate}
    \item \href{https://cs.stanford.edu/people/jcjohns/clevr/}{CLEVR}
    \item \href{http://ptr.csail.mit.edu/}{PTR} 
    \item \href{https://cs.stanford.edu/people/dorarad/gqa/index.html}{GQA} 
\end{enumerate}

The experiment code can be found in the \textit{code} folder provided along with the supplemental submission. The folder structure is provided in the \textit{README.md} in the root folder and separate files are provided to process the dataset as well as run each experiment for the different model families on different datasets. 

\subsection{Additional Figures and analysis}

Figure \ref{fig:GPT-4-LLM-v-multimodal-QUESTIONS} shows the performance of GPT-4 versus GPT-4 V on the question categories using standard prompting on the 2 datasets, while Figure \ref{fig:CoT-QUESTION} provides the same comparison for GPT-3.5-Turbo.

\begin{figure*}[htbp]
  \centering
    \includegraphics[width=\textwidth]{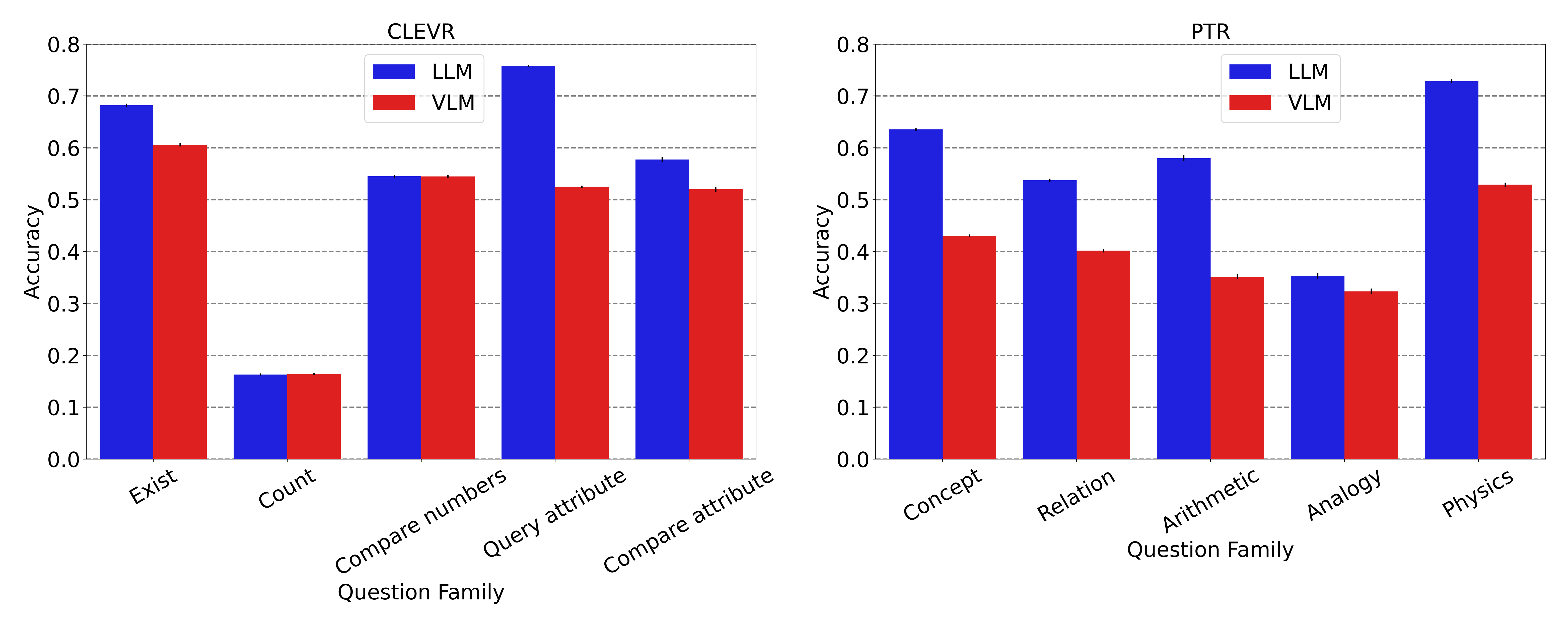}
  \caption{LLM versus VLM model performance of GPT-4 on CLEVR and PTR using standard prompting, organized by question family.}
  \label{fig:GPT-4-LLM-v-multimodal-QUESTIONS}
\end{figure*}

\begin{figure*}[htbp]
  \centering
    \includegraphics[width=\textwidth]{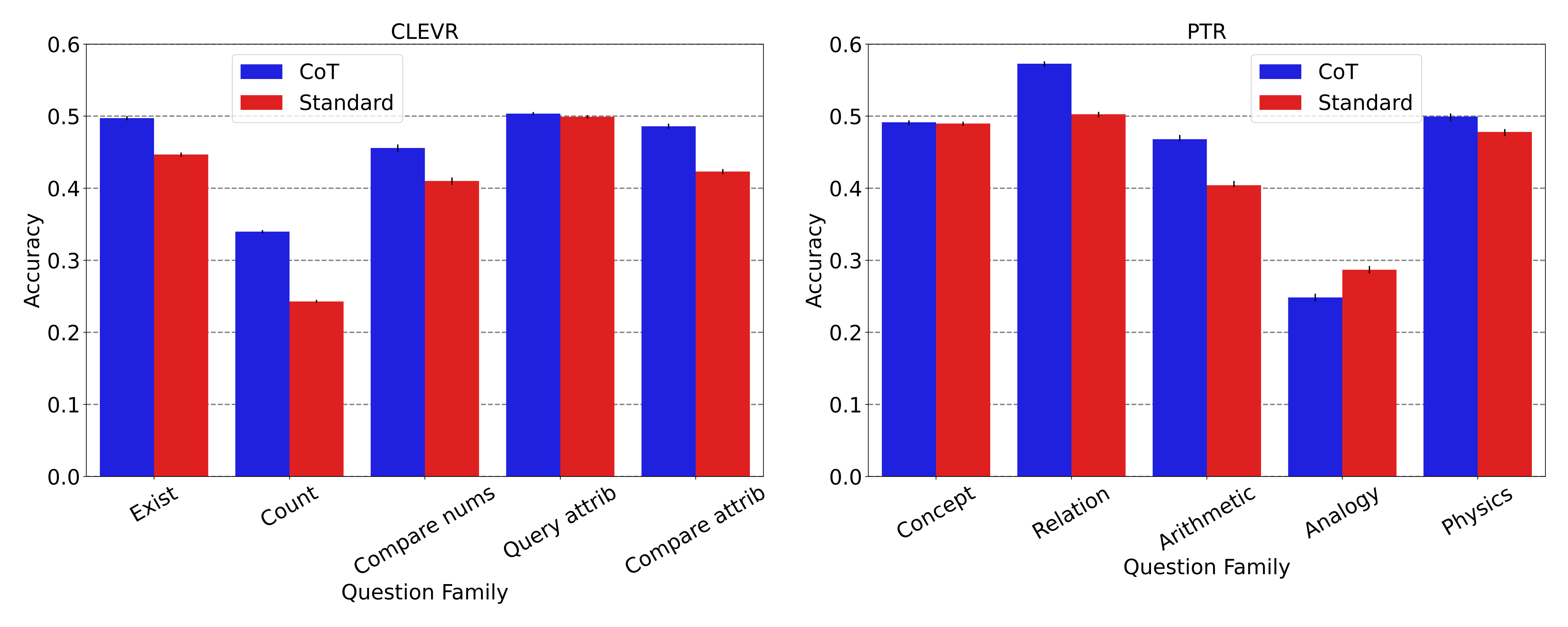}
  \caption{Standard versus CoT prompting performance of GPT-3.5-Turbo on CLEVR and PTR.}
  \label{fig:CoT-QUESTION}
\end{figure*}

\subsection{Full Prompt Examples}
\label{prompt-examples}

\subsection{CLEVR Prompt Example}
\subsubsection{\textbf{Image}}

\begin{figure*}[htbp]
    \centering
    \includegraphics[width=0.8\linewidth]{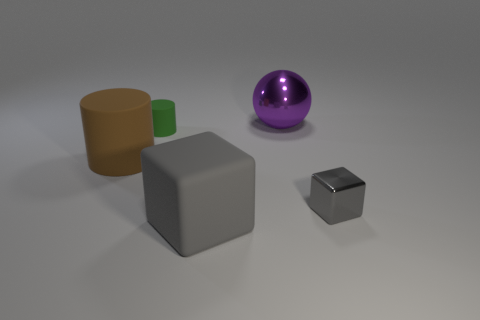}
    \caption{Example CLEVR Image used provided to the Multimodal Models.}
    \label{fig:clevr-img}
\end{figure*}

The example image used to demonstrate the prompting is provided in figure \ref{fig:clevr-img}.

\subsubsection{\textbf{Standard Prompt}}
\label{ex:cot-std}
Given the following scene:

Scene 0:

  Objects: 5
    Object:
      Color: brown
      Size: large
      Rotation: 178.92387258999463
      Shape: cylinder
      Material: rubber
      3D Coords: [-1.4937210083007812, -1.9936031103134155, 0.699999988079071]
      Pixel Coords: [119, 131, 10.801968574523926]

    Object:
      Color: gray
      Size: large
      Rotation: 243.405459279722
      Shape: cube
      Material: rubber
      3D Coords: [1.555708646774292, -2.104736566543579, 0.699999988079071]
      Pixel Coords: [198, 190, 8.60103988647461]

    Object:
      Color: green
      Size: small
      Rotation: 230.45235024165092
      Shape: cylinder
      Material: rubber
      3D Coords: [-2.342184543609619, -0.5205014944076538, 0.3499999940395355]
      Pixel Coords: [161, 118, 12.372727394104004]

    Object:
      Color: purple
      Size: large
      Rotation: 31.654351858799153
      Shape: sphere
      Material: metal
      3D Coords: [-0.8073106408119202, 1.914123773574829, 0.699999988079071]
      Pixel Coords: [282, 100, 12.495001792907715]

    Object:
      Color: gray
      Size: small
      Rotation: 42.183287560575
      Shape: cube
      Material: metal
      3D Coords: [2.6763813495635986, 0.03453871235251427, 0.3499999940395355]
      Pixel Coords: [337, 195, 9.161211967468262]

  Relationships: {'right': [[1, 2, 3, 4], [3, 4], [1, 3, 4], [4], []], 'behind': [[2, 3], [0, 2, 3, 4], [3], [], [0, 2, 3]], 'front': [[1, 4], [], [0, 1, 4], [0, 1, 2, 4], [1]], 'left': [[], [0, 2], [0], [0, 1, 2], [0, 1, 2, 3]]}

  Directions: {'right': [0.6563112735748291, 0.7544902563095093, -0.0], 'behind': [-0.754490315914154, 0.6563112735748291, 0.0], 'above': [0.0, 0.0, 1.0], 'below': [-0.0, -0.0, -1.0], 'left': [-0.6563112735748291, -0.7544902563095093, 0.0], 'front': [0.754490315914154, -0.6563112735748291, -0.0]}

  Image Filename: CLEVR\_val\_000000.png
  
You may assume that any metal object is shiny, and any rubber object is not shiny ("matte"). All objects are either "metal" or "rubber", and in 2 sizes: "large" or "small". All objects are one of the following colours: "blue", "brown", "cyan", "gray", "green", "purple", "red", "yellow". All objects are one of the following shapes: "cube", "cylinder", "sphere". For numeric answers, give an integer and not in words.

Now answer the following question in one word.

Question: Are there any other things that are the same shape as the big metallic object?
Answer: 

\subsubsection{\textbf{Chain-of-Thought Prompt}}
\label{ex:cot-clevr}
Given the following scene:

Scene 0:

  Objects: 5
    Object:
      Color: brown
      Size: large
      Rotation: 178.92387258999463
      Shape: cylinder
      Material: rubber
      3D Coords: [-1.4937210083007812, -1.9936031103134155, 0.699999988079071]
      Pixel Coords: [119, 131, 10.801968574523926]

    Object:
      Color: gray
      Size: large
      Rotation: 243.405459279722
      Shape: cube
      Material: rubber
      3D Coords: [1.555708646774292, -2.104736566543579, 0.699999988079071]
      Pixel Coords: [198, 190, 8.60103988647461]

    Object:
      Color: green
      Size: small
      Rotation: 230.45235024165092
      Shape: cylinder
      Material: rubber
      3D Coords: [-2.342184543609619, -0.5205014944076538, 0.3499999940395355]
      Pixel Coords: [161, 118, 12.372727394104004]

    Object:
      Color: purple
      Size: large
      Rotation: 31.654351858799153
      Shape: sphere
      Material: metal
      3D Coords: [-0.8073106408119202, 1.914123773574829, 0.699999988079071]
      Pixel Coords: [282, 100, 12.495001792907715]

    Object:
      Color: gray
      Size: small
      Rotation: 42.183287560575
      Shape: cube
      Material: metal
      3D Coords: [2.6763813495635986, 0.03453871235251427, 0.3499999940395355]
      Pixel Coords: [337, 195, 9.161211967468262]

  Relationships: {'right': [[1, 2, 3, 4], [3, 4], [1, 3, 4], [4], []], 'behind': [[2, 3], [0, 2, 3, 4], [3], [], [0, 2, 3]], 'front': [[1, 4], [], [0, 1, 4], [0, 1, 2, 4], [1]], 'left': [[], [0, 2], [0], [0, 1, 2], [0, 1, 2, 3]]}

  Directions: {'right': [0.6563112735748291, 0.7544902563095093, -0.0], 'behind': [-0.754490315914154, 0.6563112735748291, 0.0], 'above': [0.0, 0.0, 1.0], 'below': [-0.0, -0.0, -1.0], 'left': [-0.6563112735748291, -0.7544902563095093, 0.0], 'front': [0.754490315914154, -0.6563112735748291, -0.0]}

  Image Filename: CLEVR\_val\_000000.png

You may assume that any metal object is shiny, and any rubber object is not shiny ("matte"). All objects are either "metal" or "rubber", and in 2 sizes: "large" or "small". All objects are one of the following colours: "blue", "brown", "cyan", "gray", "green", "purple", "red", "yellow". All objects are one of the following shapes: "cube", "cylinder", "sphere". For numeric answers, give an integer and not in words. 

Now answer the following questions with step-by-step reasoning. Give the final one word answer at the end of your reasoning. Thus, your response format should be:

Reasoning for the answer \\
Final answer: \\
Final one word answer 

Question: Are there any other things that are the same shape as the big metallic object?
Answer: Let's think step by step.

\subsection{PTR Prompt Example}
\subsubsection{\textbf{Image}}
\begin{figure*}[htbp]
    \centering
    \includegraphics[width=0.8\linewidth]{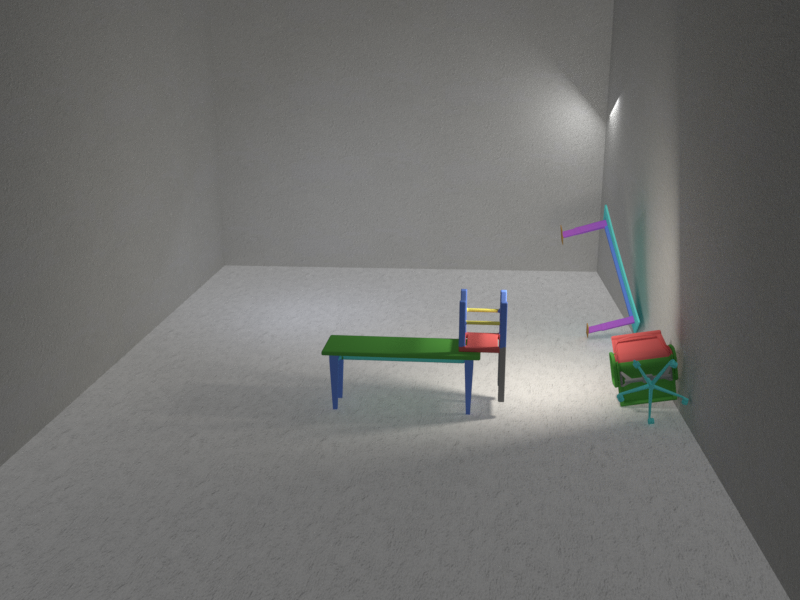}
    \caption{Example PTR Image used provided to the Multimodal Models.}
    \label{fig:ptr-img}
\end{figure*}

The example image used to demonstrate the prompting is provided in figure \ref{fig:ptr-img}.

\subsubsection{\textbf{Standard Prompt}}
\label{ex:ptr-std}
Given the following scene:

Scene PTR\_val\_007239:

  Objects: 4
    Object:
      Category: Chair
      Rotation: [3.1370301246643066, 0.17649400234222412, 3.115612745285034]
      Scale: 1.0257009267807007
      Stability: no
      3D Coords: [4.433284759521484, -6.149937629699707, 0.6772643327713013]
      Support: [645, 369, 14.225968360900879]
      Part Colors: {'arm': ['green', [29, 105, 20]], 'back': ['red', [173, 35, 35]], 'central support': ['cyan', [41, 208, 208]], 'leg': ['cyan', [41, 208, 208]], 'seat': ['green', [29, 105, 20]], 'wheel': ['cyan', [41, 208, 208]]}
      Part Count: {'leg': 5, 'wheel': 5}

    Object:
      Category: Table
      Rotation: [1.5707963705062866, -0.0, 3.115298271179199]
      Scale: 0.7057468096415201
      Stability: yes
      3D Coords: [0.4326867163181305, -6.359785556793213, 0.8900529742240906]
      Support: [398, 355, 14.242145538330078]
      Part Colors: {'leg': ['blue', [42, 75, 215]], 'top': ['green', [29, 105, 20]]}
      Part Count: {'leg': 3}

    Object:
      Category: Chair
      Rotation: [-1.5707963705062866, -0.0, 1.5445020198822021]
      Scale: 1.0358330011367798
      Stability: no
      3D Coords: [1.8846343755722046, -6.353758335113525, 1.1621549129486084]
      Support: [489, 341, 14.07287311553955]
      Part Colors: {'back': ['gray', [87, 87, 87]], 'leg': ['blue', [42, 75, 215]], 'leg bar': ['yellow', [255, 238, 51]], 'seat': ['red', [173, 35, 35]]}
      Part Count: {'leg': 4, 'leg bar': 2}

    Object:
      Category: Table
      Rotation: [2.6077208518981934, -1.005200743675232, 2.0337021350860596]
      Scale: 0.6892068386077881
      Stability: no
      3D Coords: [4.483290195465088, -2.520329236984253, 1.323003888130188]
      Support: [614, 275, 17.525205612182617]
      Part Colors: {'leg': ['purple', [129, 38, 192]], 'leg bar': ['brown', [129, 74, 25]], 'top': ['cyan', [41, 208, 208]]}
      Part Count: {'leg': 2, 'leg bar': 2}

  Relationships: {'above': [[], [2], [], []], 'behind': [[3], [3], [3], []], 'below': [[], [], [1], []], 'front': [[], [], [], [0, 1, 2]], 'left': [[1, 2], [], [], [0, 1, 2]], 'right': [[3], [0, 3], [0, 3], []]}

  Directions: {'above': [0.0, 0.0, 1.0], 'behind': [-0.05208918824791908, 0.9986424446105957, 0.0], 'below': [-0.0, -0.0, -1.0], 'front': [0.05208918824791908, -0.9986424446105957, -0.0], 'left': [-0.9986424446105957, -0.05208919197320938, 0.0], 'right': [0.9986424446105957, 0.05208919197320938, -0.0]}

  Image Filename: PTR\_val\_007239.png

  Physics: True

  Cam location: [1.4220809936523438, -19.768001556396484, 5.674197196960449]

  Cam Rotation: [0.7979968190193176, 0.6015914678573608, 0.02161088027060032, 0.028666317462921143]
The objects or things can have the following categories: 'Bed', 'Cart', 'Chair', 'Refrigerator', 'Table'. The different parts of the things can have the following categories: arm', 'arm horizontal bar', 'arm vertical bar', 'back', 'behind', 'body', 'central support', 'door', 'drawer', 'leg', 'leg bar', 'pedestal', 'seat', 'shelf', 'sleep area', 'top', 'wheel'. The things or objects can move in the following directions to make themselves stable: 'front', 'left', 'right'. The objects or their parts can have the following colors: 'blue', 'brown', 'cyan', 'gray', 'green', 'purple', 'red', 'yellow'. For numeric answers, give an answer in integers and not in words.

Now answer the following question in one word.

Question: how many objects are stable? Answer:

\subsubsection{\textbf{Chain-of-Thought Prompt}}
\label{ex;ptr-cot}
Given the following scene:

Scene PTR\_val\_007239:

  Objects: 4
    Object:
      Category: Chair
      Rotation: [3.1370301246643066, 0.17649400234222412, 3.115612745285034]
      Scale: 1.0257009267807007
      Stability: no
      3D Coords: [4.433284759521484, -6.149937629699707, 0.6772643327713013]
      Support: [645, 369, 14.225968360900879]
      Part Colors: {'arm': ['green', [29, 105, 20]], 'back': ['red', [173, 35, 35]], 'central support': ['cyan', [41, 208, 208]], 'leg': ['cyan', [41, 208, 208]], 'seat': ['green', [29, 105, 20]], 'wheel': ['cyan', [41, 208, 208]]}
      Part Count: {'leg': 5, 'wheel': 5}

    Object:
      Category: Table
      Rotation: [1.5707963705062866, -0.0, 3.115298271179199]
      Scale: 0.7057468096415201
      Stability: yes
      3D Coords: [0.4326867163181305, -6.359785556793213, 0.8900529742240906]
      Support: [398, 355, 14.242145538330078]
      Part Colors: {'leg': ['blue', [42, 75, 215]], 'top': ['green', [29, 105, 20]]}
      Part Count: {'leg': 3}

    Object:
      Category: Chair
      Rotation: [-1.5707963705062866, -0.0, 1.5445020198822021]
      Scale: 1.0358330011367798
      Stability: no
      3D Coords: [1.8846343755722046, -6.353758335113525, 1.1621549129486084]
      Support: [489, 341, 14.07287311553955]
      Part Colors: {'back': ['gray', [87, 87, 87]], 'leg': ['blue', [42, 75, 215]], 'leg bar': ['yellow', [255, 238, 51]], 'seat': ['red', [173, 35, 35]]}
      Part Count: {'leg': 4, 'leg bar': 2}

    Object:
      Category: Table
      Rotation: [2.6077208518981934, -1.005200743675232, 2.0337021350860596]
      Scale: 0.6892068386077881
      Stability: no
      3D Coords: [4.483290195465088, -2.520329236984253, 1.323003888130188]
      Support: [614, 275, 17.525205612182617]
      Part Colors: {'leg': ['purple', [129, 38, 192]], 'leg bar': ['brown', [129, 74, 25]], 'top': ['cyan', [41, 208, 208]]}
      Part Count: {'leg': 2, 'leg bar': 2}

  Relationships: {'above': [[], [2], [], []], 'behind': [[3], [3], [3], []], 'below': [[], [], [1], []], 'front': [[], [], [], [0, 1, 2]], 'left': [[1, 2], [], [], [0, 1, 2]], 'right': [[3], [0, 3], [0, 3], []]}

  Directions: {'above': [0.0, 0.0, 1.0], 'behind': [-0.05208918824791908, 0.9986424446105957, 0.0], 'below': [-0.0, -0.0, -1.0], 'front': [0.05208918824791908, -0.9986424446105957, -0.0], 'left': [-0.9986424446105957, -0.05208919197320938, 0.0], 'right': [0.9986424446105957, 0.05208919197320938, -0.0]}

  Image Filename: PTR\_val\_007239.png

  Physics: True

  Cam location: [1.4220809936523438, -19.768001556396484, 5.674197196960449]

  Cam Rotation: [0.7979968190193176, 0.6015914678573608, 0.02161088027060032, 0.028666317462921143]
The objects or things can have the following categories: 'Bed', 'Cart', 'Chair', 'Refrigerator', 'Table'. The different parts of the things can have the following categories: arm', 'arm horizontal bar', 'arm vertical bar', 'back', 'behind', 'body', 'central support', 'door', 'drawer', 'leg', 'leg bar', 'pedestal', 'seat', 'shelf', 'sleep area', 'top', 'wheel'. The things or objects can move in the following directions to make themselves stable: 'front', 'left', 'right'. The objects or their parts can have the following colors: 'blue', 'brown', 'cyan', 'gray', 'green', 'purple', 'red', 'yellow'. For numeric answers, give an answer in integers and not in words.

Now answer the following questions with step-by-step reasoning.Give the final one word answer at the end of your reasoning. Thus, your response format should be:

Reasoning for the answer \\
Final answer: \\
Final one word answer

Question: how many objects are stable? Answer: Let's think step by step.

\subsection{Prompt For VLM with full scene metadata}
In the case of VLMs with full scene metadata, we use the same exact scene description as provided for Standard prompts, with the image input used the same as in VLM experiments.  

\subsection{Examples of Reasoning step complexities}

More details about the the reasoning steps and question families can be found within the papers of the respective datasets:

- CLEVR Dataset [Figure 2 of the paper \cite{johnson2017clevr}]

- PTR Dataset [Figure 1 of the paper \cite{hong2021ptr}

\subsection{Example of VLM vs LLM}

\begin{figure*}[htbp]
  \centering
    \includegraphics[width=\textwidth]{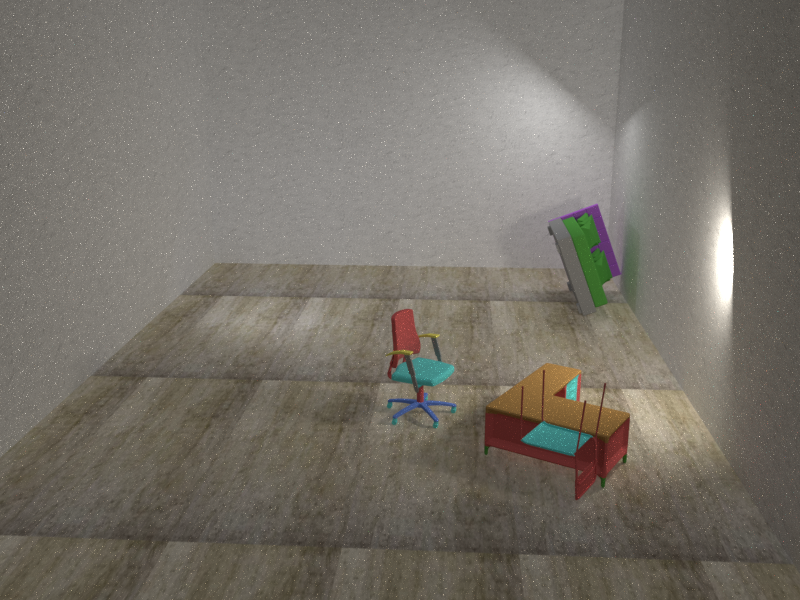}
  \caption{Example of a PTR scene.}
    \label{fig:PTR-val-example}
\end{figure*}

\begin{figure*}[htbp]
  \centering
    \includegraphics[width=\textwidth]{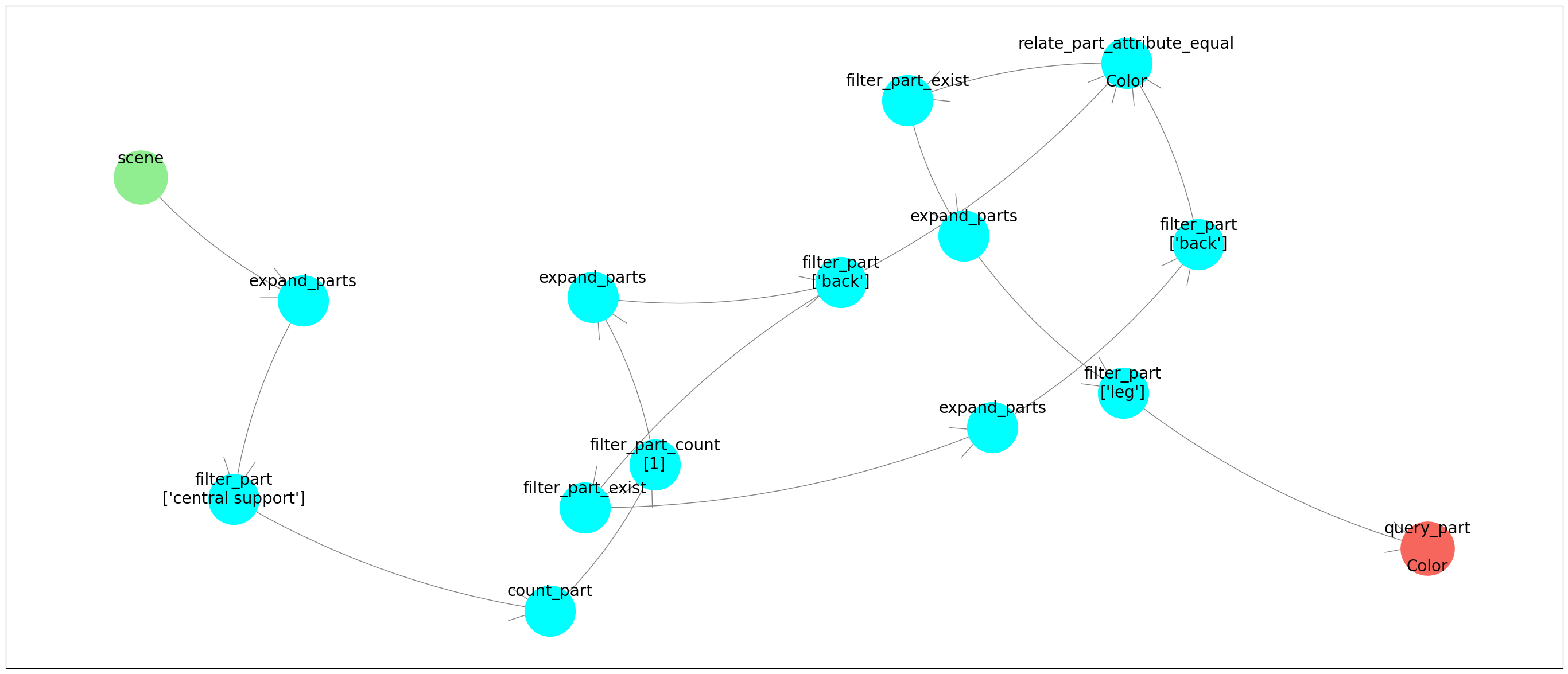}
  \caption{Reasoning steps required at the answer for the question ``what is the color of the legs of the thing that has the same color of back as the object with a central support?''. The Green Node signifies the input step while the Red node signifies the output step. The arrows indicate the flow of reasoning.}
    \label{fig:PTR-val-reasoning-steps}
\end{figure*}

Figure \ref{fig:PTR-val-example} shows a scene used in the evaluation of the GPT-4 and GPT-4 Vision models from the PTR dataset. The following was one of the questions asked to the models:  ``what is the color of the legs of the thing that has the same color of back as the object with a central support?'' 

Figure \ref{fig:PTR-val-reasoning-steps} shows the reasoning steps required to reach the answer starting from the scene input. The correct answer for this question was ``red''. The following were the answers provided by the LLM and the VLM:

- GPT-4 (LLM + scene metadata): ``red''

- GPT-4 Vision (VLM): ``cyan''

As we can see, the LLM arrives at the answer correctly, while the VLM fails to do so. This example concretely shows the complex reasoning required to arrive at the answer, as well as a case where the LLM performs better at reasoning than the VLM.

\subsection{Full Experimental Results}
\label{experimental-results}
The results for all experiments performed are given in the Table \ref{tab:results}

\begin{table*}[htbp]
\centering
\caption{Experiment Results}
\scriptsize
\label{tab:results}
\begin{tabular}{|c|c|c|p{2cm}|p{3cm}|c|}
\toprule
  Model &  Scale (Billions of parameters) & Dataset &       Type & Prompting Technique &  Accuracy \\
\midrule
FLAN T5 &   3.00 &   CLEVR &        LLM &                 CoT &  0.250172 \\
BLIP-2 FLAN T5 &   3.00 &   CLEVR & Multimodal &                 CoT &  0.138455 \\
FLAN T5 &   3.00 &   CLEVR &        LLM &            Standard &  0.463932 \\
BLIP-2 FLAN T5 &   3.00 &   CLEVR & Multimodal &            Standard &  0.396497 \\
BLIP-2 FLAN T5 &  3.00 &   CLEVR & Multimodal + Full Scene Metadata &            Standard &  0.455474 \\
FLAN T5 &  11.00 &   CLEVR &        LLM &                 CoT &  0.360632 \\
BLIP-2 FLAN T5 &  11.00 &   CLEVR & Multimodal &                 CoT &  0.282964 \\
FLAN T5 &  11.00 &   CLEVR &        LLM &            Standard &  0.463932 \\
BLIP-2 FLAN T5 &  11.00 &   CLEVR & Multimodal &            Standard &  0.402938 \\
BLIP-2 FLAN T5 &  11.00 &   CLEVR & Multimodal + Full Scene Metadata &            Standard &  0.481456 \\
    GPT &   0.35 &   CLEVR &        LLM &                 CoT &  0.085992 \\
    GPT &   0.35 &   CLEVR &        LLM &            Standard &  0.095729 \\
    GPT &   1.30 &   CLEVR &        LLM &                 CoT &  0.127561 \\
    GPT &   1.30 &   CLEVR &        LLM &            Standard &  0.175713 \\
    GPT &   6.70 &   CLEVR &        LLM &                 CoT &  0.124974 \\
    GPT &   6.70 &   CLEVR &        LLM &            Standard &  0.296915 \\
    GPT & 175.00 &   CLEVR &        LLM &                 CoT &  0.457613 \\
    GPT & 175.00 &   CLEVR &        LLM &            Standard &  0.409037 \\
    GPT & 1800.00 &   CLEVR &        LLM &            Standard &  0.55625 \\
    GPT & 1800.00 &   CLEVR &        Multimodal + Full Scene Metadata &            Standard &  0.584012 \\
    GPT & 1800.00 &   CLEVR &        Multimodal &            Standard &  0.450969 \\
    GEMINI & N/A &   CLEVR &        LLM &            Standard &  0.528713 \\
    GEMINI & N/A &   CLEVR &        Multimodal &            Standard &  0.403052 \\
    GEMINI & N/A &   CLEVR &        Multimodal + Full Scene Metadata &            Standard &  0.496572 \\
    \midrule
FLAN T5 &   3.00 &     PTR &        LLM &                 CoT &  0.349455 \\
BLIP-2 FLAN T5 &   3.00 &     PTR & Multimodal &                 CoT &  0.071239 \\
FLAN T5 &   3.00 &     PTR &        LLM &            Standard &  0.508657 \\
BLIP-2 FLAN T5 &   3.00 &     PTR & Multimodal &            Standard &  0.336524 \\
BLIP-2 FLAN T5 &  3.00 &   PTR & Multimodal + Full Scene Metadata &            Standard &  0.488672 \\
FLAN T5 &  11.00 &     PTR &        LLM &                 CoT &  0.375339 \\
BLIP-2 FLAN T5 &  11.00 &     PTR & Multimodal &                 CoT &  0.228783 \\
FLAN T5 &  11.00 &     PTR &        LLM &            Standard &  0.531447 \\
BLIP-2 FLAN T5 &  11.00 &     PTR & Multimodal &            Standard &  0.352028 \\
BLIP-2 FLAN T5 &  11.00 &   PTR & Multimodal + Full Scene Metadata &            Standard &  0.522143 \\
    GPT &   0.35 &     PTR &        LLM &                 CoT &  0.036834 \\
    GPT &   0.35 &     PTR &        LLM &            Standard &  0.038419 \\
    GPT &   1.30 &     PTR &        LLM &                 CoT &  0.053044 \\
    GPT &   1.30 &     PTR &        LLM &            Standard &  0.149849 \\
    GPT &   6.70 &     PTR &        LLM &                 CoT &  0.057316 \\
    GPT &   6.70 &     PTR &        LLM &            Standard &  0.242263 \\
    GPT & 175.00 &     PTR &        LLM &                 CoT &  0.486693 \\
    GPT & 175.00 &     PTR &        LLM &            Standard &  0.461586 \\
    GPT & 175.00 &     PTR &        LLM &                 CoT &  0.486693 \\
    GPT & 175.00 &     PTR &        LLM &            Standard &  0.461586 \\
    GPT & 1800.00 &   PTR &        LLM &            Standard &  0.586388 \\
    GPT & 1800.00 &   PTR &        Multimodal + Full Scene Metadata &            Standard &  0.601769 \\
    GPT & 1800.00 &   PTR &        Multimodal &            Standard &  0.409766 \\
    GEMINI & N/A &   PTR &        LLM &            Standard &  0.550529 \\
    GEMINI & N/A &   PTR &        Multimodal &            Standard &  0.391480 \\
    GEMINI & N/A &   PTR &        Multimodal + Full Scene Metadata &            Standard &  0.481256 \\
    
\bottomrule
\end{tabular}
\end{table*}

\subsubsection{VLM CoT Performance Discussion}

The decision to omit the VLM CoT results from the main paper was mainly due to space constraints and our initial assessment that these results might not offer as much insight as the other findings. We aimed to streamline the main content of the paper for clarity and conciseness. 

Here are the results and discussion for the VLM CoT Prompting performance:

\textbf{Observations}

For both the CLEVR and PTR datasets, the accuracy of BLIP-2 Flan-T5 XXL models is generally higher than the BLIP-2 Flan-T5 XL models, regardless of the prompting technique. Across both datasets and both model sizes, the Standard Prompting technique consistently outperforms the CoT (Chain of Thought) prompting technique. The performance drop due to CoT is more pronounced in the smaller BLIP-2 Flan-T5 XL models compared to the larger XXL variants. For example, in the CLEVR dataset, the XL model shows a drop of 25.8\% (from 39.65\% to 13.85\%) when using CoT compared to Standard Prompting, while the XXL variant shows a smaller drop of 12\% (from 40.29\% to 28.29\%). A similar trend is observed in the PTR dataset where the performance drop in the XL model is 26.53\% (from 33.65\% to 7.12\%) compared to a drop of 12.32\% (from 35.20\% to 22.88\%) in the XXL model when switching from Standard Prompting to CoT.

\textbf{Analysis}

\textbf{Prompting Technique Influence.} While Standard Prompting seems to be the more effective method across both datasets, the Chain of Thought (CoT) reasoning does show potential, and trends of emergence over scale. This is especially important considering that the VLMs are not explicitly trained on synthetic images, suggesting that CoT emergence in VLMs is not limited to the tasks or image categories for which they were trained. Additionally, it provides further evidence for the observation that CoT does seem to emerge even in the absence of world knowledge.

\textbf{Implications for Future Research.} While scale evidently improves performance, there's a need to further investigate the interaction between prompting techniques and model scale. The larger drop in performance in smaller models when using CoT indicates that certain reasoning capabilities emerge more robustly at higher scales. Future research could delve deeper into optimizing prompting techniques specifically for smaller models or further enhancing the performance of larger models.

\subsection{GQA Experiments}
\label{GQA-results}

The GQA dataset was used to test the experimental setup on a dataset which uses natural images instead of synthetically generated images. This was done in order to check the fairness of the VLM vs LLM comparision on the original datasets. The rationale behind this was that the Visual encoders in the VLMs were not trained on synthetic images, which affect the performance on the datasets selected in the original paper. The GQA dataset was as it provided access to comprehensive scene metadata as well as functional programs to arrive at the answer, similar to the \citep{johnson2017clevr} and \citep{hong2021ptr} datasets used in the main experiments. 
To facilitate our analysis, we used annotated scene graphs as a proxy for perfect scene information. Further, since GQA contains 1469 labels which can hinder a model’s effectiveness (when providing the label vocabulary through prompting), we drew a subset of questions for the top-25 labels and used a maximum prompt length of 20,000 tokens. This resulted in 78536 questions out of the original 132062 questions in the official validation set. The code to sample the dataset and run the experiments is provided along with the supplemental submission \ref{code-details}. 
we observe that the LLM (Flan-T5-XXL) has accuracy of 78.72\%  while the VLM (BLIP2-Flan-T5-XXL) with the same base language model has accuracy of 56.81\%. The overall model performance is provided in Table~\ref{tab:GQA_results}, performance over length of functional programs is provided in Table~\ref{tab:length_performance} and the performance over the question families is provided in Table~\ref{tab:question_family_performance}.

\textbf{Analysis of the results.} We can see that the LLM outperforms the VLM on the dataset, as well as over the length of functional programs and question families. This result is consistent with the findings of the main paper. It is important to note that there are not many questions with a large length of the functional programs in the dataset, the scene metadata covers all the important relationships and informations in a more verbose manner and the answers seem to be generally simpler to answer than the synthetic datasets, which could explain a relatively larger gap in the LLM vs VLM performance.

\begin{table}[ht]
\centering
\caption{Experiment Results on Sampled GQA Dataset}
\label{tab:GQA_results}
\begin{tabular}{|c|c|c|}
\toprule
  Model           & Dataset              & Accuracy \\
\midrule
Flan-T5 XXL       & Sampled GQA Dataset  & 78.72 \\
Blip-2 Flan-T5 XXL & Sampled GQA Dataset  & 56.81 \\
\bottomrule
\end{tabular}
\end{table}

\begin{table}[ht]
\centering
\caption{Performance Over Length of Functional Programs on GQA}
\label{tab:length_performance}
\begin{tabular}{|c|c|c|c|c|c|}
\toprule
Length & Model           & Correct & Total & Accuracy (\%) \\
\midrule
2 & Flan-T5 XXL       & 20404 & 26977 & 75.63 \\
2 & Blip-2 Flan-T5 XXL & 14517 & 26977 & 53.81 \\
3 & Flan-T5 XXL       & 23921 & 29574 & 80.88 \\
3 & Blip-2 Flan-T5 XXL & 16914 & 29574 & 57.19 \\
4 & Flan-T5 XXL       & 6685 & 8100 & 82.53 \\
4 & Blip-2 Flan-T5 XXL & 4920 & 8100 & 60.74 \\
5 & Flan-T5 XXL       & 7819 & 10369 & 75.40 \\
5 & Blip-2 Flan-T5 XXL & 6538 & 10369 & 63.05 \\
6 & Flan-T5 XXL       & 77   & 83   & 92.77 \\
6 & Blip-2 Flan-T5 XXL & 56   & 83   & 67.47 \\
7 & Flan-T5 XXL       & 2908 & 3426 & 84.88 \\
7 & Blip-2 Flan-T5 XXL & 1664 & 3426 & 48.57 \\
8 & Flan-T5 XXL       & 6    & 6    & 100.0    \\
8 & Blip-2 Flan-T5 XXL & 5    & 6    & 83.33    \\
9 & Flan-T5 XXL       & 1    & 1    & 100.0     \\
9 & Blip-2 Flan-T5 XXL & 1    & 1    & 100.0   \\
\bottomrule
\end{tabular}
\end{table}

\begin{table*}[htbp]
\centering
\caption{Performance Over Question Families on GQA}
\label{tab:question_family_performance}
\begin{tabular}{|c|c|c|c|c|}
\toprule
Question Family & Model            & Correct & Total & Accuracy (\%) \\
\midrule
Logical & Flan T5 XXL          & 12503 & 15590 & 80.19 \\
Logical & Blip-2 Flan T5 XXL   & 9829 & 15590 & 59.58 \\
Verify  & Flan T5 XXL          & 21145& 26355 & 80.23 \\
Verify  & Blip-2 Flan T5 XXL   & 15967 & 26355 & 60.58 \\
Query   & Flan T5 XXL          & 18027 & 21910 & 82.27 \\
Query   & Blip-2 Flan T5 XXL   & 12669 & 21910 & 57.82 \\
Choose  & Flan T5 XXL          & 7585 & 11374 & 66.68 \\
Choose  & Blip-2 Flan T5 XXL   & 4925 & 11374 & 45.30 \\
Compare & Flan T5 XXL          & 2561 & 3307 & 77.44 \\
Compare & Blip-2 Flan T5 XXL   & 1765  & 3307  & 53.27 \\
\bottomrule
\end{tabular}
\end{table*}

\subsubsection{GQA Example Prompt}

\begin{figure*}[htbp]
    \centering
    \includegraphics[width=0.8\linewidth]{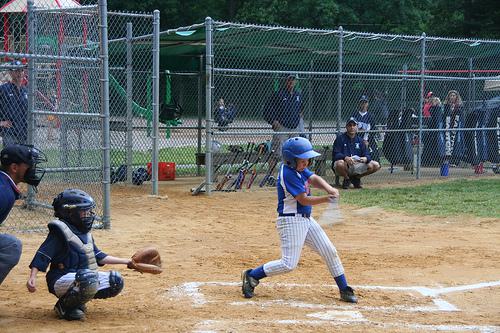}
    \caption{Example GQA Image used provided to the VLM Models.}
    \label{fig:gqa-img}
\end{figure*}

\textbf{Image}
The example image used to demonstrate the prompting for GQA is provided in \ref{fig:gqa-img}

\textbf{Standard Prompt}

context: Given the following scene:
Image Dimensions: 500x347
Objects: 34
  Object ID 1231798:
    Name: face
    Coordinates: x=442, y=55
    Dimensions: w=16, h=25
    Attributes: 
      Relation:
        Name: of
        Object: 1231760
      Relation:
        Name: to the right of
        Object: 1231777
      Relation:
        Name: to the right of
        Object: 1231778

  Object ID 1231799:
    Name: face
    Coordinates: x=68, y=20
    Dimensions: w=25, h=36
    Attributes: 
      Relation:
        Name: to the right of
        Object: 1231788
      Relation:
        Name: of
        Object: 1231770
      Relation:
        Name: to the left of
        Object: 1231781
      Relation:
        Name: to the left of
        Object: 1231773

  Object ID 1231790:
    Name: cart
    Coordinates: x=168, y=141
    Dimensions: w=168, h=192
    Attributes: 
      Relation:
        Name: to the right of
        Object: 1231781
      Relation:
        Name: to the right of
        Object: 1231780
      Relation:
        Name: to the right of
        Object: 1231770
      Relation:
        Name: to the right of
        Object: 1231779
      Relation:
        Name: to the right of
        Object: 1231797
      Relation:
        Name: to the right of
        Object: 1231796
      Relation:
        Name: to the left of
        Object: 1231760
      Relation:
        Name: to the right of
        Object: 1231768
      Relation:
        Name: to the right of
        Object: 1231767
      Relation:
        Name: to the left of
        Object: 1231763

  Object ID 1231792:
    Name: outlet
    Coordinates: x=199, y=338
    Dimensions: w=13, h=7
    Attributes: 
      Relation:
        Name: on
        Object: 1231793

  Object ID 1231793:
    Name: floor
    Coordinates: x=0, y=167
    Dimensions: w=498, h=180
    Attributes:

  Object ID 1231794:
    Name: arm
    Coordinates: x=43, y=76
    Dimensions: w=70, h=48
    Attributes: 
      Relation:
        Name: to the left of
        Object: 1231781
      Relation:
        Name: to the right of
        Object: 1231786
      Relation:
        Name: to the left of
        Object: 1231802
      Relation:
        Name: to the left of
        Object: 1231783

  Object ID 1231796:
    Name: leg
    Coordinates: x=96, y=184
    Dimensions: w=31, h=104
    Attributes: 
      Relation:
        Name: of
        Object: 1231770
      Relation:
        Name: to the left of
        Object: 1231781
      Relation:
        Name: to the right of
        Object: 1231797
      Relation:
        Name: to the left of
        Object: 1231790
      Relation:
        Name: to the left of
        Object: 1231767
      Relation:
        Name: to the left of
        Object: 1231766

  Object ID 1231797:
    Name: leg
    Coordinates: x=13, y=187
    Dimensions: w=56, h=123
    Attributes: 
      Relation:
        Name: to the left of
        Object: 1231796
      Relation:
        Name: to the left of
        Object: 1231768
      Relation:
        Name: of
        Object: 1231770
      Relation:
        Name: to the left of
        Object: 1231767
      Relation:
        Name: to the left of
        Object: 1231766
      Relation:
        Name: to the left of
        Object: 1231781
      Relation:
        Name: to the left of
        Object: 1231790

  Object ID 1231800:
    Name: shirt
    Coordinates: x=20, y=56
    Dimensions: w=95, h=136
    Attributes: 
      Relation:
        Name: to the left of
        Object: 1231802
      Relation:
        Name: to the left of
        Object: 1231781
      Relation:
        Name: to the left of
        Object: 1231764
      Relation:
        Name: to the left of
        Object: 1231779
      Relation:
        Name: to the right of
        Object: 1231786

  Object ID 1231802:
    Name: shirt
    Coordinates: x=113, y=57
    Dimensions: w=60, h=93
    Attributes: 
      Relation:
        Name: to the left of
        Object: 1231783
      Relation:
        Name: to the left of
        Object: 1231764
      Relation:
        Name: to the right of
        Object: 1231800
      Relation:
        Name: to the right of
        Object: 1231794
      Relation:
        Name: to the right of
        Object: 1231786
      Relation:
        Name: to the left of
        Object: 1231782
      Relation:
        Name: to the left of
        Object: 1231777
      Relation:
        Name: to the left of
        Object: 1231778

  Object ID 1231769:
    Name: sandal
    Coordinates: x=2, y=304
    Dimensions: w=31, h=38
    Attributes: brown

  Object ID 1231768:
    Name: sandal
    Coordinates: x=96, y=286
    Dimensions: w=51, h=26
    Attributes: brown
      Relation:
        Name: to the left of
        Object: 1231790
      Relation:
        Name: to the right of
        Object: 1231797

  Object ID 1231761:
    Name: arm
    Coordinates: x=430, y=85
    Dimensions: w=40, h=64
    Attributes: 
      Relation:
        Name: to the right of
        Object: 1231762
      Relation:
        Name: to the right of
        Object: 1231782
      Relation:
        Name: to the right of
        Object: 1231777

  Object ID 1231760:
    Name: girl
    Coordinates: x=370, y=48
    Dimensions: w=107, h=224
    Attributes: 
      Relation:
        Name: to the right of
        Object: 1231777
      Relation:
        Name: carrying
        Object: 1231762
      Relation:
        Name: to the right of
        Object: 1231790
      Relation:
        Name: to the right of
        Object: 1231764
      Relation:
        Name: to the right of
        Object: 1231783
      Relation:
        Name: to the right of
        Object: 1231782
      Relation:
        Name: with
        Object: 1231762
      Relation:
        Name: to the right of
        Object: 1231778

  Object ID 1231763:
    Name: sneakers
    Coordinates: x=427, y=251
    Dimensions: w=27, h=20
    Attributes: white
      Relation:
        Name: to the right of
        Object: 1231790

  Object ID 1231762:
    Name: purse
    Coordinates: x=392, y=114
    Dimensions: w=52, h=55
    Attributes: 
      Relation:
        Name: to the right of
        Object: 1231777
      Relation:
        Name: to the right of
        Object: 1231764
      Relation:
        Name: to the right of
        Object: 1231782
      Relation:
        Name: to the left of
        Object: 1231761

  Object ID 1231764:
    Name: projector
    Coordinates: x=196, y=126
    Dimensions: w=74, h=30
    Attributes: 
      Relation:
        Name: to the right of
        Object: 1231800
      Relation:
        Name: to the right of
        Object: 1231802
      Relation:
        Name: to the left of
        Object: 1231760
      Relation:
        Name: to the left of
        Object: 1231762
      Relation:
        Name: to the left of
        Object: 1231782
      Relation:
        Name: to the right of
        Object: 1231781
      Relation:
        Name: to the left of
        Object: 1231777
      Relation:
        Name: to the right of
        Object: 1231770
      Relation:
        Name: to the right of
        Object: 1231779

  Object ID 1231767:
    Name: sandal
    Coordinates: x=134, y=246
    Dimensions: w=31, h=19
    Attributes: brown
      Relation:
        Name: to the left of
        Object: 1231790
      Relation:
        Name: to the right of
        Object: 1231770
      Relation:
        Name: to the right of
        Object: 1231797
      Relation:
        Name: to the right of
        Object: 1231796

  Object ID 1231766:
    Name: sandal
    Coordinates: x=164, y=237
    Dimensions: w=39, h=12
    Attributes: brown
      Relation:
        Name: to the right of
        Object: 1231780
      Relation:
        Name: to the right of
        Object: 1231797
      Relation:
        Name: to the right of
        Object: 1231796
      Relation:
        Name: to the right of
        Object: 1231770

  Object ID 1231788:
    Name: books
    Coordinates: x=41, y=21
    Dimensions: w=9, h=19
    Attributes: 
      Relation:
        Name: to the left of
        Object: 1231799
      Relation:
        Name: to the left of
        Object: 1231771
      Relation:
        Name: to the left of
        Object: 1231773
      Relation:
        Name: to the left of
        Object: 1231781

  Object ID 1231783:
    Name: wall
    Coordinates: x=223, y=0
    Dimensions: w=72, h=120
    Attributes: brick
      Relation:
        Name: to the right of
        Object: 1231781
      Relation:
        Name: to the left of
        Object: 1231760
      Relation:
        Name: to the left of
        Object: 1231777
      Relation:
        Name: to the right of
        Object: 1231787
      Relation:
        Name: to the right of
        Object: 1231802
      Relation:
        Name: to the right of
        Object: 1231794
      Relation:
        Name: to the right of
        Object: 1231770
      Relation:
        Name: to the left of
        Object: 1231778
      Relation:
        Name: to the right of
        Object: 1231773

  Object ID 1231782:
    Name: game
    Coordinates: x=268, y=135
    Dimensions: w=40, h=18
    Attributes: 
      Relation:
        Name: to the right of
        Object: 1231770
      Relation:
        Name: to the right of
        Object: 1231779
      Relation:
        Name: to the left of
        Object: 1231762
      Relation:
        Name: to the right of
        Object: 1231764
      Relation:
        Name: to the right of
        Object: 1231781
      Relation:
        Name: to the right of
        Object: 1231802
      Relation:
        Name: to the left of
        Object: 1231761
      Relation:
        Name: to the left of
        Object: 1231760

  Object ID 1231781:
    Name: people
    Coordinates: x=112, y=24
    Dimensions: w=88, h=240
    Attributes: 
      Relation:
        Name: to the right of
        Object: 1231788
      Relation:
        Name: to the left of
        Object: 1231783
      Relation:
        Name: to the left of
        Object: 1231764
      Relation:
        Name: to the right of
        Object: 1231787
      Relation:
        Name: to the right of
        Object: 1231786
      Relation:
        Name: to the left of
        Object: 1231782
      Relation:
        Name: to the right of
        Object: 1231780
      Relation:
        Name: to the left of
        Object: 1231777
      Relation:
        Name: in front of
        Object: 1231786
      Relation:
        Name: wearing
        Object: 1231779
      Relation:
        Name: to the right of
        Object: 1231774
      Relation:
        Name: to the left of
        Object: 1231778
      Relation:
        Name: to the right of
        Object: 1231770
      Relation:
        Name: playing
        Object: 1231782
      Relation:
        Name: to the right of
        Object: 1231794
      Relation:
        Name: to the left of
        Object: 1231790
      Relation:
        Name: to the right of
        Object: 1231799
      Relation:
        Name: to the right of
        Object: 1231800
      Relation:
        Name: to the right of
        Object: 1231797
      Relation:
        Name: to the right of
        Object: 1231796
      Relation:
        Name: wearing
        Object: 1231780
      Relation:
        Name: wearing
        Object: 1231802

  Object ID 1231780:
    Name: shorts
    Coordinates: x=22, y=175
    Dimensions: w=106, h=74
    Attributes: 
      Relation:
        Name: to the left of
        Object: 1231779
      Relation:
        Name: to the left of
        Object: 1231766
      Relation:
        Name: to the left of
        Object: 1231790
      Relation:
        Name: to the left of
        Object: 1231781

  Object ID 1231787:
    Name: shelf
    Coordinates: x=16, y=7
    Dimensions: w=97, h=39
    Attributes: brown
      Relation:
        Name: to the left of
        Object: 1231781
      Relation:
        Name: to the left of
        Object: 1231783
      Relation:
        Name: by
        Object: 1231786
      Relation:
        Name: to the left of
        Object: 1231773

  Object ID 1231786:
    Name: window
    Coordinates: x=1, y=76
    Dimensions: w=20, h=58
    Attributes: 
      Relation:
        Name: to the left of
        Object: 1231781
      Relation:
        Name: behind
        Object: 1231781
      Relation:
        Name: to the left of
        Object: 1231794
      Relation:
        Name: to the left of
        Object: 1231802
      Relation:
        Name: to the left of
        Object: 1231800

  Object ID 1231776:
    Name: hair
    Coordinates: x=432, y=49
    Dimensions: w=22, h=17
    Attributes: 
      Relation:
        Name: to the right of
        Object: 1231778
      Relation:
        Name: to the right of
        Object: 1231777

  Object ID 1231777:
    Name: girl
    Coordinates: x=274, y=42
    Dimensions: w=60, h=194
    Attributes: 
      Relation:
        Name: to the right of
        Object: 1231802
      Relation:
        Name: to the right of
        Object: 1231764
      Relation:
        Name: to the left of
        Object: 1231762
      Relation:
        Name: to the left of
        Object: 1231760
      Relation:
        Name: to the right of
        Object: 1231783
      Relation:
        Name: to the right of
        Object: 1231781
      Relation:
        Name: to the left of
        Object: 1231776
      Relation:
        Name: to the right of
        Object: 1231779
      Relation:
        Name: to the left of
        Object: 1231798
      Relation:
        Name: to the left of
        Object: 1231761

  Object ID 1231774:
    Name: beard
    Coordinates: x=73, y=49
    Dimensions: w=17, h=12
    Attributes: 
      Relation:
        Name: to the left of
        Object: 1231781
      Relation:
        Name: to the left of
        Object: 1231773

  Object ID 1231773:
    Name: hair
    Coordinates: x=124, y=25
    Dimensions: w=30, h=35
    Attributes: 
      Relation:
        Name: to the right of
        Object: 1231774
      Relation:
        Name: to the right of
        Object: 1231799
      Relation:
        Name: to the left of
        Object: 1231783
      Relation:
        Name: to the right of
        Object: 1231787
      Relation:
        Name: to the right of
        Object: 1231788

  Object ID 1231770:
    Name: man
    Coordinates: x=3, y=7
    Dimensions: w=145, h=333
    Attributes: 
      Relation:
        Name: to the left of
        Object: 1231779
      Relation:
        Name: to the left of
        Object: 1231764
      Relation:
        Name: to the left of
        Object: 1231790
      Relation:
        Name: wearing
        Object: 1231762
      Relation:
        Name: to the left of
        Object: 1231766
      Relation:
        Name: to the left of
        Object: 1231767
      Relation:
        Name: wearing
        Object: 1231800
      Relation:
        Name: wearing
        Object: 1231780
      Relation:
        Name: to the left of
        Object: 1231781
      Relation:
        Name: to the left of
        Object: 1231782
      Relation:
        Name: to the left of
        Object: 1231783
      Relation:
        Name: with
        Object: 1231774

  Object ID 1231771:
    Name: hair
    Coordinates: x=48, y=7
    Dimensions: w=48, h=33
    Attributes: 
      Relation:
        Name: to the right of
        Object: 1231788

  Object ID 1231778:
    Name: hair
    Coordinates: x=286, y=42
    Dimensions: w=38, h=43
    Attributes: 
      Relation:
        Name: to the right of
        Object: 1231802
      Relation:
        Name: to the left of
        Object: 1231760
      Relation:
        Name: to the right of
        Object: 1231783
      Relation:
        Name: to the right of
        Object: 1231781
      Relation:
        Name: to the left of
        Object: 1231798
      Relation:
        Name: to the left of
        Object: 1231776

  Object ID 1231779:
    Name: shorts
    Coordinates: x=119, y=133
    Dimensions: w=59, h=83
    Attributes: 
      Relation:
        Name: to the left of
        Object: 1231790
      Relation:
        Name: to the right of
        Object: 1231770
      Relation:
        Name: to the right of
        Object: 1231800
      Relation:
        Name: to the left of
        Object: 1231764
      Relation:
        Name: to the right of
        Object: 1231780
      Relation:
        Name: to the left of
        Object: 1231782
      Relation:
        Name: to the left of
        Object: 1231777

 The possible answers could be: yes, no, left, right, man, white, black, bottom, woman, blue, chair, top, brown, table, boy, gray, bed, green, girl, red, cat, dog, car, bus, horse.Now answer the following question in one word.
Question: What color is the helmet in the middle of the image?

\subsection{Image-Free Baseline and Random Chance}
\label{image_free}

We conducted image-free experiments for GPT 3.5 models on both CLEVR and PTR to establish a baseline to which the model performance could be compared. This meant providing the~\textbf{Dataset name, dataset split (val) and image name,} followed by \textbf{answer vocabulary hint} to the model, similar to the original prompt, and then asking the question, \textbf{without providing any scene metadata}. Since the GPT 3.5 model was trained on the open internet, there is a chance that it could have seen some of the dataset during its training process. Establishing an image-free baseline enabled us to gauge whether the model had prior information about the questions and the scenes. 

For Flan-T5, the datasets on which the model was trained have been disclosed and do not contain CLEVR or PTR. More details about Flan-T5 training and fine-tuning is available in Appendix F of the main paper.

The code required to run the experiments have been provided with the supplemental submission under the "image\_free" folder

\textbf{Image-free baseline prompt -- CLEVR}

Answer the following question from the val split of the CLEVR Dataset for image CLEVR\_val\_000000.png
You may assume that any metal object is shiny, and any rubber object is not shiny ("matte"). All objects are either "metal" or "rubber", and in 2 sizes: "large" or "small". All objects are one of the following colours: "blue", "brown", "cyan", "gray", "green", "purple", "red", "yellow". All objects are one of the following shapes: "cube", "cylinder", "sphere". For numeric answers, give an integer and not in words.
Always answer the following question in a single word from the options provided above. Your response should only be a single word.
Question: Is there a big brown object of the same shape as the green thing?

Answer:

\textbf{Image-free baseline prompt -- PTR}

Answer the following question from the val split of the PTR Dataset for image PTR\_val\_007239.png
The objects or things can have the following categories: 'Bed', 'Cart', 'Chair', 'Refrigerator', 'Table'. The different parts of the things can have the following categories: arm', 'arm horizontal bar', 'arm vertical bar', 'back', 'behind', 'body', 'central support', 'door', 'drawer', 'leg', 'leg bar', 'pedestal', 'seat', 'shelf', 'sleep area', 'top', 'wheel'. The things or objects can move in the following directions to make themselves stable: 'front', 'left', 'right'. The objects or their parts can have the following colors: 'blue', 'brown', 'cyan', 'gray', 'green', 'purple', 'red', 'yellow'. For numeric answers, give an answer in integers and not in words.
Always answer the following question in a single word from the options provided above. Your response should be just a single word.
Question : how many objects are stable?

Answer:

\textbf{Image-free baseline -- Results}

The models response for most such questions would be that "\textbf{The question cannot be answered without more information}". Thus, we forced a valid response from the model by asking it to always give a one word answer from the answer vocabulary provided. 

\textit{CLEVR Image-free Baseline}

The Image free baseline performance of GPT 3.5 on CLEVR was \textbf{36.85\%}. Table \ref{tab:random_chance_clevr} calculates the random chance of getting a question from CLEVR right. We see that the image free baseline results indicate that the model performance in the absence of scene metadata is basically random chance. 

\begin{table}[ht]
\centering
\caption{Random Chance and Total Questions for CLEVR}
\label{tab:random_chance_clevr}
\begin{tabular}{|l|c|c|}
\toprule
Category & Random Chance (\%) & Total Questions \\
\midrule
exist               & 50.00 & 20196 \\
colors              & 12.50 & 13404 \\
material            & 50.00 & 30545 \\
compare attribute   & 50.00 & 35422 \\
shape               & 33.33 & 13544 \\
size                & 50.00 & 10094 \\
count               & 10.00 & 13273 \\
compare numbers     & 50.00 & 13513 \\
\midrule
\multicolumn{2}{|l|}{Overall random chance} & 36.86 \\
\bottomrule
\end{tabular}
\end{table}

\textit{PTR Image-free Baseline}

The Image free baseline performance of GPT 3.5 on CLEVR was \textbf{10.16\%}. Table \ref{tab:random_chance_ptr} calculates the random chance of getting a question from CLEVR right. Again, we see that the image free baseline results indicate that the model performance in the absence of scene metadata is basically random chance.

\begin{table}[ht]
\centering
\caption{Random Chance and Total Questions for PTR}
\label{tab:random_chance_ptr}
\begin{tabular}{|l|c|c|}
\toprule
Category & Random Chance (\%) & Total Questions \\
\midrule
concept     & 2.63  & 38972 \\
relation    & 4.35  & 22905 \\
physics     & 50.00 & 7413 \\
analogy     & 5.26  & 7472 \\
arithmetic  & 8.33  & 14958 \\
\midrule
\multicolumn{2}{|l|}{Overall random chance} & 8.03 \\
\bottomrule
\end{tabular}
\end{table}

\subsection{\textbf{Compute Used}} 
The models were trained using different compute resources depending on the scale. All Multimodal models were run on NVIDIA A100 (40GB) VRAM GPUs. For the FLAN-T5 family, the 11B models used NVIDIA A100 (40GB), while the smaller ones (3B) used NVIDIA A40's. Complete details can be found in \ref{tab:compute}.

\begin{table*}[htbp]
\centering
\caption{Compute Used for Different Models}
\label{tab:compute}
\begin{tabular}{|c|c|c|c|}
\toprule
\textbf{Model} & \textbf{Compute Used} & \textbf{Experiment (Both CLEVR and PTR)}\\
\midrule
BLIP-2 FLAN-T5 (3B) & NVIDIA A40 & Multimodal CoT \\
BLIP-2 FLAN-T5 (3B) & NVIDIA A40 & Multimodal Standard\\
FLAN-T5 (3B) & NVIDIA A40 & LLM CoT\\
FLAN-T5 (3B) & NVIDIA A40 & LLM Standard \\
BLIP-2 FLAN-T5 (3B) & NVIDIA A40 & Multimodal CoT \\
BLIP-2 FLAN-T5 (3B) & NVIDIA A40 & Multimodal Standard  \\
FLAN-T5 (3B) & NVIDIA A40 & LLM CoT \\
FLAN-T5 (3B) & NVIDIA A40 & LLM Standard \\
BLIP-2 FLAN-T5 (11B) & NVIDIA A100 (40GB) & Multimodal CoT\\
BLIP-2 FLAN-T5 (11B) & NVIDIA A100 (40GB) & Multimodal Standard \\
FLAN-T5 (11B) & NVIDIA A100 (40GB) & LLM CoT\\
FLAN-T5 (11B) & NVIDIA A100 (40GB) & LLM Standard\\
BLIP-2 FLAN-T5 (11B) & NVIDIA A100 (40GB) & Multimodal CoT\\
BLIP-2 FLAN-T5 (11B) & NVIDIA A100 (40GB) & Multimodal Standard\\
FLAN-T5 (11B) & NVIDIA A100 (40GB) & LLM CoT \\
FLAN-T5 (11B) & NVIDIA A100 (40GB) & LLM Standard \\
Gemini & Gemini API & LLM Standard \\
Gemini & Gemini Vision API & Multimodal Standard \\
GPT (1.7T) & OpenAI API & LLM CoT \\
GPT (1.7T) & OpenAI API & LLM Standard \\
GPT (1.7T) & OpenAI API & Multimodal CoT\\
GPT (1.7T) & OpenAI API & Multimodal Standard \\
GPT (175B) & OpenAI API & LLM CoT \\
GPT (175B) & OpenAI API & LLM Standard \\
GPT (350M) & OpenAI API & LLM CoT \\
GPT (350M) & OpenAI API & LLM Standard \\
GPT (1.3B) & OpenAI API & LLM CoT  \\
GPT (1.3B) & OpenAI API & LLM Standard\\
GPT (6.7B) & OpenAI API & LLM CoT \\
GPT (6.7B) & OpenAI API & LLM Standard \\
\bottomrule
\end{tabular}
\end{table*}

\subsection{Examples of limitations}
\subsubsection{Gibberish Answers}

We will use the full example for the CLEVR dataset provided in \ref{ex:cot-clevr} to demonstrate some commonly types of non-sensical answers which the models provide. 

\textbf{Repeating the Question}
The following question was provided to the Flan-T5 3B parameter LLM model with the chain-of-thought prompt provided in section \ref{ex:cot-clevr} as scene description. 

\textit{Question}: There is a small gray block; are there any spheres to the left of it?

\textit{Model Response}:  There is a small gray block; are there any spheres to the right of it?

Here the model just repeats the question provided to it.

\textbf{Repeating the input prompt}
The following question was provided to the Flan-T5 3B parameter LLM model with the chain-of-thought prompt provided in section \ref{ex:cot-clevr} as scene description. 

\textit{Question}: What is the color of the large shiny sphere?

\textit{Model Response}:  You may assume that any metal object is shiny, and any rubber object is not shiny ("matte"). All objects are either metallic or rubbery, and in 2 sizes: "large" or "small". All objects are one of the following colours: "blue", "brown", "cyan", "gray", "green", "purple", "red", "yellow". All objects are one of the following shapes: "cube", "cylinder", "sphere". For numeric answers, give an integer and not in words. 

Here the model simply repeats the setup prompt which was used to align its responses with the dataset answer vocabulary.

\textbf{Repeating the same logic over and over}

\textit{Question}: Are there any other things that are the same shape as the big metallic object?

\textit{Model Response}:  The big metallic object is a cube. The small metallic object is a cube. The purple object is a cube. The red object is a cube. The yellow object is a cube. The blue object is a cube. The brown object is a cube. The green object is a cube. The red object is a cube. The yellow object is a cube. The blue object is a cube. The brown object is a cube. The purple object is a cube. The red object is a cube. The yellow object is a cube. The blue object is a cube. The brown object is a cube. The green object is a cube. The purple object is a cube. The red object is a cube. The yellow object is a cube. The blue object is a cube. The brown object is a cube. The purple object is a cube. The yellow object is a cube. The red object is a cube. The brown object is a cube. The green object is a cube. The yellow object is a cube. The blue object is a cube. The brown object is a cube. The purple object is a cube. The yellow object is a cube. The blue object is a cube. The brown object is a cube. The purple object is a cube. The yellow object is a cube. The blue object is a cube. The brown object is a cube. The purple object is a cube. The yellow object is a cube. The blue object is a cube. The brown object is a cube. The purple object is a cube. The yellow object is a cube. The blue object is a cube. The brown object is a cube. The purple object is a cube. The yellow object is a cube. The blue object is a cube. The brown object is a cube. The purple object is a cube. The yellow object is a cube. The blue object is a cube. The brown object is a cube. The purple object

As seen here, the model gets lost in its own explanations and keeps repeating the same logic over and over. 

\textbf{Not adhere to the response format}
The following question was provided to the Flan-T5   3B parameter LLM model with the chain-of-thought prompt provided in section \ref{ex:cot-clevr} as scene description. 

\textit{Question}: Is there a big brown object of the same shape as the green thing?

\textit{Model Response}:  The green object is a cube, and the brown object is a cube too

As seen here, while the model response is correct, it does not follow the response format provided to it. 

\subsubsection{Prompt Engineering}

Prompt engineering plays an important role in eliciting the required reasoning behaviour from the instruction based LLMs used in the project. Changes in the wording of the prompt, or emphasis on different aspects of the instructions can have drastic changes in the model response. 

\textbf{Guidance for Response Structure}. We noticed the most impact in model responses on the answer formatting aspect of the prompt. The larger scale models were significantly better in following a required format for their response, while the smaller models required further emphasis for the same. For example, the smaller model variants of both the GPT as well as Flan-T5   models adhered to the required format much better when the response format was repeated at the end of the prompt as well. 

\textbf{Impact of prompt format}. The formatting of the input prompt affects the model response as well. The Flan-T5 models   are text-to-text models which have been trained on question answering in a specific format. During their training, they were made to elicit CoT reasoning for their responses by providing samples in the following format:

\underline{Example of the ideal prompt} \\
$<$Scene metadata + setup prompt$>$ \\
\textit{Question}: $<$Question about the provided scene$>$ \\
\textit{Answer}: Lets think step-by-step.

From our observations, the Flan-T5 models were able to provide best responses for CoT reasoning when provided the prompt in this exact format. Other formats for the prompts, such as those given below,  were not as effective in making the models perform CoT reasoning. 

\underline{Examples of ineffective prompts}

\textit{Example 1} \\
$<$Scene metadata + setup prompt$>$ \\
Answer the following question with step-by-step reasoning: \\
\textit{Question}: $<$Question about the provided scene$>$ \\
\textit{Answer}: 

\textit{Example 2} \\
$<$Scene metadata + setup prompt$>$ \\
Answer the following question with step-by-step reasoning: \\
\textit{Question}: $<$Question about the provided scene$>$ \\
Lets think step-by-step.

This indicates that while the models can exhibit general reasoning abilities, their performance is not yet flexible or robust to prompts they might not have observed in their training data before. Their performance for the same task still heavily depends on the format and content of the training data. 

\textbf{Observations}: From empirical observations, it seems like the models become more robust to prompt format and better at following instructions with scale, since the models of the same type show progressive improvements in both these aspects as scale increases.

\end{document}